\documentclass{ps}
\usepackage{amsthm,amsmath}
\RequirePackage[numbers]{natbib}
\RequirePackage[colorlinks,citecolor=blue,urlcolor=blue]{hyperref}
\usepackage{amssymb}
\usepackage{color}
\usepackage{hyperref}
\usepackage{comment}
\usepackage{mathrsfs}
\usepackage{algorithm}
\usepackage[noend]{algpseudocode}
\usepackage{wasysym}
\usepackage{graphicx}
\usepackage{bbm}
\usepackage[warning,easyold,math]{easyeqn}
\usepackage{changepage}

\def\argmax{\operatornamewithlimits{argmax}}
\def\cond{\, \big| \,}

\renewcommand\leq{\leqslant}
\renewcommand\geq{\geqslant}

\newcommand\E{\mathbb E}

\newcommand\p{\mathbb P}

\newcommand\D{\mathcal D}
\newcommand\x{\mathcal X}
\newcommand\y{\mathcal Y}

\newcommand\Be{\text{Bernoulli}}

\newcommand\R{\mathbb R}
\newcommand\N{\mathbb N}

\newcommand\II{\mathbbm 1}

\newcommand\tstar{\theta^*}
\newcommand\that{\widehat\theta}
\newcommand\ttilde{\widetilde\theta}

\newcommand\thati[1]{\widehat\theta^{(#1)}}
\newcommand\ttildei[1]{\widetilde\theta^{(#1)}}
\newcommand\tlinei[1]{\overline\theta^{(#1)}}

\newcommand\etilde{\widetilde\eta}
\newcommand\eline{\overline\eta}

\newcommand\etildei[1]{\widetilde\eta^{(#1)}}
\newcommand\elinei[1]{\overline\eta^{(#1)}}

\newcommand\kl[2]{\mathcal{K}\left( #1, #2 \right)}
\newcommand\klhalf[2]{\mathcal{K}^{1/2}\left( #1, #2 \right)}

\begin{document}
\title{An adaptive multiclass nearest neighbor classifier}
\thanks{Financial support by the Russian Academic Excellence Project 5-100
	and by the German Research Foundation (DFG) through the Collaborative 
	Research Center 1294 is gratefully acknowledged.}
\runningtitle{An adaptive multiclass nearest neighbor classifier}
\author{Nikita Puchkin}
\address{National Research University Higher School of Economics, 20 
Myasnitskaya ulitsa, 101000, Moscow, RF, 
\href{mailto:npuchkin@hse.ru}{npuchkin@hse.ru} }
\secondaddress{Institute for Information Transmission Problems RAS, Bolshoy Karetny 
per. 19, 127051, Moscow, RF.}
\author{Vladimir Spokoiny}
\address{Weierstrass Institute and Humboldt University, Mohrenstr. 39, 10117 
Berlin, Germany}
\sameaddress{, 2, 1}
\runningauthors{N. Puchkin and V. Spokoiny}
%
%
\begin{abstract}
	We consider a problem of multiclass classification, where the training sample 
	\break $S_n =  \{(X_i, Y_i)\}_{i=1}^n$ is generated from the model $\p(Y = m | 
	X = x) = \eta_m(x)$, $1 \leq m \leq M$, and $\eta_1(x), \dots, \eta_M(x)$ are 
	unknown $\alpha$-Holder continuous functions.
	Given a test point $X$, our goal is to predict its label.
	A widely used $\mathsf k$-nearest-neighbors classifier constructs estimates of 
	$\eta_1(X), \dots, \eta_M(X)$ and uses a plug-in rule for the prediction.
	However, it requires a proper choice of the smoothing parameter $\mathsf k$, which 
	may become tricky in some situations.
	We fix several integers $n_1, \dots, n_K$, compute corresponding 
	$n_k$-nearest-neighbor estimates for each $m$ and each $n_k$ and apply an 
	aggregation procedure.
	We study an algorithm, which constructs a convex combination of these estimates 
	such that the aggregated estimate behaves approximately as well as an oracle 
	choice.
	We also provide a non-asymptotic analysis of the procedure, prove its adaptation to 
	the unknown smoothness parameter $\alpha$ and to the margin and establish rates 
	of convergence under mild assumptions.	
\end{abstract}
\begin{resume}
Nous consid\'erons un probl\'eme de classification multi-classes, dans lequel 
l'\'echantillon \break d'apprentissage $S_n =  \{(X_i, Y_i)\}_{i=1}^n$ est engendr\'e par le 
model 
$\p(Y = m | X = x) = \eta_m(x)$, $1 \leq m \leq M$, o\'u les fonctions $\eta_1(x), \dots, 
\eta_M(x)$ sont inconnues et $\alpha$-H\"{o}ld\'eriennes.
Etant donn\'e un point test $X$, notre but est de pr\'edire sont label $Y$.
Dans ce contexte, une m\'ethode fr\'equemment utilis\'ee consiste \'a produire des 
estimateurs $\eta_1(X), \dots, \eta_M(X)$ de type $k$-plus proches voisins et de les 
combiner par une m\'ethode par insertion.
Cependant, cette approche suppose un choix judicieux du param\'etre de r\'egularization 
$k$ qui peut \^etre parfois difficile.
Dans notre approche, nous fixons des entiers $n_1, \dots, n_K$, calculons l'estimateur 
des $n_k$-plus proches voisins pour tout $m$ et tout $n_k$ et utilisons une proc\'edure 
d'agr\'egation.
Nous \'etudions un algorithme qui construit une combinaison convexe de ces 
estimateurs et dont la performance est comparable \'a celle d'un choix oracle.
Nous proposons aussi une \'etude non-asymptotique de cette m\'ethode, prouvons 
qu'elle s'adapte \'a la r\'egularit\'e inconnue $\alpha$ ainsi qu'\'a la marge et 
\'etablissons 
des vitesses de convergence sous des hypoth\'eses  faibles.
\end{resume}
\subjclass{62H30, 62G08}
\keywords{multiclass classification, $\mathsf k$ nearest neighbors, adaptive 
procedures}
\maketitle

\section*{Introduction}

Multiclass classification is a natural generalization of the well-studied problem of binary 
classification.
It is a problem of supervised learning when one observes a sample \( S_n = \left\{ (X_1, 
Y_1), \dots, (X_n, Y_n) \right\} \), where \( X_i \in \x \subseteq \R^d \), \( Y_i \in \y = \{ 1, 
\dots, M \} \), \( 1 \leq i \leq n \), \( M > 2 \).
The pairs \( (X_i, Y_i) \) are generated independently according to an unknown 
distribution \( \D \) over \( \x \times \y \). 
Given a test pair $(X, Y)$, which is generated from $\D$ independently of $S_n$, the 
learner's task is to propose a rule \( f : \mathcal X \rightarrow \{ 1, \dots, M \} \) in order to 
make a probability of misclassification
\[
R(f) = \p_{(X, Y) \sim \D} \left( Y \neq f(X) \right)
\]
as small as possible.
In practice, it is a common situation when one has to discern between more than two 
classes, so multiclass classification has a wide range of applications and arises in such 
areas as bioinformatics, when one tries to predict a protein's fold \cite{protein} or when 
one wants to classify DNA microarrays \cite{cancer}, finance when one predicts a 
corporate credit rating \cite{credit_rating}, image analysis \cite{face_verification} when 
one tries to classify an object on an image, speech recognition \cite{speech} and others.

Concerning the multiclass learning problem, one can distinguish between two main 
approaches.
The first one is by reducing to the binary classification.
The most popular and straightforward examples of these techniques are One-vs-All 
(OvA) and One-vs-One (OvO).
Another example of reduction to the binary case is given by error correcting output 
codes (ECOC) \cite{db95}.
In \cite{ass01}, this approach was generalized for margin classifiers.
A similar approach uses tree-based classifiers.
Methods of the second type solve a single problem such as it is done in 
multiclass SVM \cite{cs02} and multiclass one-inclusion graph strategy \cite{rbr06}.
Daniely, Sabato, and Shalev-Shwartz in \cite{dss12} provided a theoretical comparison 
of  OvA, OvO, ECOC, tree-based classifiers and multiclass SVM for linear discrimination 
rules in a finite-dimensional space.
From their study, multiclass SVM outperforms the OvA method. 
In \cite{cs02}, Crammer and Singer also showed a superiority of multiclass SVM on 
several datasets.
Nevertheless, in our work, we will use One-vs-All for two reasons.
First, we will consider a broad nonparametric class of functions and results in 
\cite{dss12} do not cover this case.
Second, in \cite{rif04}, Rifkin and Klautau showed that OvA behaves comparably 
to multiclass SVM if binary classifier in OvA is strong enough.

For each class \( m \), we construct binary labels \( \II (Y_i = m) \).
We denote a marginal distribution of $X$ by $\p_X$ and suppose that $\p_X$ has a 
density $p(X)$ with respect to the Lebesgue measure $\mu$. 
Given $X$, we denote the conditional probability $\p \left( Y = m | X \right)$, $1 \leq m 
\leq M$ by $\eta_m(X)$.
For this model, the optimal classifier \( f^* \) can be found analytically
\[
f^*(X) = \argmax\limits_{1 \leq m \leq M} \eta_m(X) .
\]
Unfortunately, true values \( \eta_1(X), \dots, \eta_M(X) \) are unknown but can be 
estimated.
Since for any classifier $f$ it holds $R(f) \geq R(f^*)$,
then it is reasonable to consider the excess risk
\[
\mathcal E(f) = R(f) - R(f^*),
\]
which shows the quantitative difference between the classifier \( f \) and the best 
possible one. 
One of the most popular approaches to tackle the classification problem is a (weighted) 
$\mathsf k$-nearest neighbors rule.
Given a test point $X \in \x$, this rule constructs nearest neighbor estimates 
$\widehat\eta_1^{(NN)}(X), \dots, \widehat\eta_M^{(NN)}(X)$ of $\eta_1(X), \dots, 
\eta_M(X)$ and predicts the label $Y$ at the point $X$ by a plug-in rule:
\[
\widehat f^{(NN)}(X) = \argmax\limits_{1 \leq m \leq M} \widehat\eta_m^{(NN)}(X),
\]
Although the method is simple and known for a long time, several new finite sample 
results in the binary setting were obtained quite recently.
In \cite{samworth12}, the author considers weighted and bagged nearest 
neighbor estimates with smooth function $\eta_1(x)$ and finds optimal vector of 
non-negative weights.
Moreover, the author goes further and derives faster rates under additional smoothness 
assumptions if the weights are allowed to be negative.
However, the analysis in \cite{samworth12} requires that the marginal distribution of 
features must have a compact support and its density must be bounded away from zero 
(strong density assumption).
In \cite{cd14} and \cite{dgw18}, authors address this issue.
In \cite{cd14}, the authors introduce a novel Holder-like smoothness condition on 
$\eta_1(x)$ tailored to nearest neighbor.
This trick allows to avoid the strong density assumption and boundedness of features.
Although in both papers \cite{cd14} and \cite{dgw18} the authors work under similar 
assumptions, their approach is different.
In \cite{cd14}, the authors obtain convergence rates in terms of size of the effective 
boundary, while in \cite{dgw18} the authors use a more classical 
approximation-estimation tradeoff technique of statistical analysis.
The disadvantage of the modified smoothness condition in \cite{cd14} is that it is 
implicit.
Instead of this condition, in \cite{gkm16}, the authors introduce the minimal mass 
assumption and the tail assumption, which are proved to be necessary for quantitative 
analysis of nearest neighbor estimates and cover the case when marginal distribution of 
features has an unbounded support and has a density, which may be arbitrarily close to 
zero.
Note that the nearest neighbor estimate $\widehat\eta_m^{(NN)}(X)$ strongly depends 
on the parameter $\mathsf k$ and its choice determines the performance of the 
classifier \( \widehat f^{(NN)} \).
Moreover, as pointed out in \cite{cbs17} and \cite{gkm16}, the global nearest neighbor 
classifier (i.e. the number of neighbors is the same for all test points) may be suboptimal, 
while the nearest neighbor classifier with point-dependent choice of $\mathsf k$ shows 
a better performance.
In multiclass setting, the situation is even more difficult, because for each class the 
optimal number of neighbors may be different and this fact becomes crucial when for 
two classes (say, 1 and 2) and a test point $X$ the values $\eta_1(X)$ and $\eta_2(X)$ 
are very close, and one has to estimate these values as precisely as possible to avoid 
misclassification.
Since for each test point $X$ and each class $m$, the optimal value of $\mathsf k$ may 
be different, the tuning procedure becomes tricky.
To solve this problem, we consider a sequence of integers $n_1, \dots, n_K$, compute 
weighted nearest neighbor estimates for each of them and use a plug-in classifier based 
on a convex combination of these estimates.

An aggregation of the nearest neighbor estimates is a key feature of our procedure.
We use a multiclass spatial stagewise aggregation (SSA), which originates from  
\cite{bs07}, where an aggregation of binary classifiers was studied.
Unlike many other aggregation procedures, such as exponential weighting 
\cite{t03, l07, exp_weighting}, mirror averaging \cite{juditsky05, jrt08}, empirical risk 
minimization \cite{l13}, and Q-aggregation \cite{dai12, lr14}, which perform 
\emph{global} aggregation, SSA makes \emph{local} aggregation yielding a point 
dependent aggregation scheme.
This means that the aggregating coefficients depend on the point \( X \) where the 
classification rule is applied.
The drawback of the original SSA procedure \cite{bs07} is that it is tightly related to the 
Kullback-Leibler aggregation and, therefore, puts some restrictions, which are usual for 
such setup and appear in other works on this topic (for instance, \cite{b16, rig12}) 
but are completely unnecessary for the classification task.
We show that, in a special case of the multiclass classification, one can omit those 
restrictions and obtain the same results under weaker assumptions.

Finally, it is worth mentioning that nonparametric estimates have slow rates of 
convergence especially in the case of high dimension \( d \).
It was shown in \cite{at07} and then in \cite{dinh14} that plug-in classifiers can achieve 
fast learning rates under certain assumptions in both binary and multiclass classification 
problems.
We will use a similar technique to derive fast learning rates for the plug-in classifier 
based on the aggregated estimate.

Main contributions of this paper are the following:
\begin{itemize}
	\item we propose a computationally efficient algorithm of multiclass classification, 
	which is based on aggregation of nearest neighbor estimates;
	\item the procedure automatically chooses an almost optimal number of neighbors 
	for each test point and each class;
	\item the procedure adapts to an unknown smoothness of the 
	functions $\eta_1(\cdot), \dots, \eta_M(\cdot)$;
	\item we provide theoretical guarantees on large deviations of the excess risk and 
	on its mean value as well under mild assumptions;
	theoretical guarantees claim optimal accuracy of classification with only a 
	logarithmic payment for adaptation.
\end{itemize}

The rest of the paper is organized as follows.
In Section \ref{notations}, we give auxiliary definitions and introduce some notations. 
In Section \ref{algorithm}, we formulate the multiclass classification procedure and
then provide its theoretical properties in Section \ref{main_results}.
Section \ref{proof} is devoted to the proof of the main result, which is given in Theorem 
\ref{main}.
Some auxiliary results and proofs are moved to the appendix. 
Finally, in Section \ref{numerical}, we demonstrate a performance of the procedure on 
both artificial and real datasets.

\section{Preliminaries and notations}
\label{notations}

We start with a simple observation.
Introduce
\begin{equation}
\label{phi}
\varphi(t) = \left( \frac1{2M} \vee t \right) \wedge \left( 1 - \frac1{2M} \right) .
\end{equation}
It is easy to show that for the truncated function
\[
\theta_m(X)
= \varphi(\eta_m(X))
\equiv \left( \frac1{2M} \vee \eta_m(X) \right) \wedge \left( 1 - \frac1{2M} \right),
\]
it holds
\[
\argmax\limits_{1 \leq m \leq M} \eta_m(X) = \argmax\limits_{1 \leq m \leq M} 
\theta_m(X),
\]
and, instead of the value $\eta_m(x)$, one can estimate $\theta_m(x)$ at a point $x$.
In our approach, we consider a plug-in classifier
\[
\label{plugin}
\widehat f(X) = \argmax\limits_{1 \leq m \leq M} \that_m(X),
\]
where \( \that_m(x) \) stands for an estimate of \( \theta_m(x) \), \( 1 \leq m \leq M 
\), at the point $x$.

Now, the problem is to estimate \( \theta_m(x) \), \( 1 \leq m \leq M \).
Fix some \( m \) and transform the labels into binarized ones: $\II \left( Y_i = m \right)$.
It is clear that
\[
\big( \II \left( Y_i = m \right) | X_i \big) \sim \Be (\eta_m(X_i)) .
\]
This approach is nothing but the One-vs-All procedure for multiclass classification.
Then a weighted $\mathsf k$-nearest-neighbor estimate of $\theta_m(x)$ at the point 
$x$ can 
be expressed as $\ttilde_m^w(x) = \varphi(\widetilde\eta_m^w(x))$ and
\begin{equation}
\label{wknn}
\widetilde\eta_m^w(x)
= \frac{\sum\limits_{i=1}^n w_i(X_i, x) \II(Y_i = m)}{\sum\limits_{i=1}^n w_i(X_i, x) }
\equiv \frac{S^w_m(x)}{N_w(x)},
\end{equation}
where \( S_m^w(x) = \sum\limits_{i=1}^n w_i(X_i, x) \II(Y_i = m) \), \( N_w(x) = 
\sum\limits_{i=1}^n w_i(X_i, x) \), is a weighted nearest neighbor estimate of $\eta_m(x)$.
The non-negative weights $w_i(X_i, x)$ depend on the distance between $X_i$ and $x$ 
and $w_i(X_i, x) > 0$ if and only if $X_i$ is among $\mathsf k$ nearest neighbors of $x$;
otherwise, $w_i(X_i, x) = 0$.
In this paper, we consider the weights of the following form:
\begin{equation}
\label{weights}
w_i = w_i(X_i, x) = \mathscr K\left( \frac{\|X_i - x\|}h \right),
\end{equation}
where a bandwidth $h = h(\mathsf k)$ is a distance to the $\mathsf k$-th nearest 
neighbor and the 
kernel $\mathscr K(\cdot)$ fulfills the following conditions:
\begin{align}
\label{a1}
& \notag
\bullet\quad
\mathscr K(t) \text{ is a non-increasing funciton},
\\& \tag{A1}
\bullet\quad
\mathscr K(0) = 1,
\\& \notag
\bullet\quad
\mathscr K(1) \geq \frac12,
\\& \notag
\bullet\quad
\mathscr K(t) = 0, \quad \forall \, t > 1.
\end{align}
This assumption can be easily satisfied.
First, note that the rectangular kernel \( \mathscr K(t) = \II \left( 0 \leq t \leq 1 \right) \) 
meets these requirements and, therefore, \eqref{a1} holds for the case of ordinary  
nearest neighbor estimates.
There are other examples of such kernels $\mathscr K$.
For instance, one can easily check that Epanechnikov-like and Gaussian-like kernels, 
$\mathscr K(t) = (1 - t^2/2)\II(0 \leq t \leq 1)$ and $\mathscr K(t) = e^{-t^2/2}\II(0 \leq t 
\leq 1)$ respectively, fulfill $\eqref{a1}$.
It is also important to mention that here and further in this paper, without loss of 
generality, we suppose that a tie (i. e. a situation, when there are several candidates for 
the $\mathsf k$-th nearest neighbor) does not happen almost surely.
Otherwise, one can use the tie-breaking procedure described in \cite{cd14}.

The nearest neighbor estimate \eqref{wknn} requires a proper choice of the parameter 
$\mathsf k$.
Moreover, an optimal value of $\mathsf k$ may be different for each test point $x$ and 
each 
class $m$, and the problem of a fine parameter tuning may become tricky.
Instead of using one universal value of the number of neighbors, we fix an increasing 
sequence of integers $\left\{ n_k : 1 \leq k \leq K \right\}$.
We only require that there exist constants $0 < u_0 < u < 1$ such that
\begin{equation}
\label{a2}
\tag{A2}
2u_0 \leq \frac{n_{k-1}}{n_k} \leq \frac u2, \quad 1 \leq k \leq K,
\end{equation}
and there are positive constants $a$ and $b$ such that $n_1 \leq a$ and $n_K \geq 
bn^{2/(d+2)}$.
Each $n_k$ induces a set of weights $w^{(k)}_1, 
\dots, w^{(k)}_n$ with
\begin{equation}
\label{weights2}
w^{(k)}_i = w^{(k)}_i(X_i, x) = \mathscr K\left( \frac{\|X_i - x\|}{h_k} \right),
\end{equation}
where $h_k$ stands for the distance to the $n_k$-th nearest neighbor, and
a weighted $n_k$-NN estimator:
\begin{equation}
\label{ttilde}
\ttildei k_m(x)
= \varphi \left( \etildei k_m(x) \right)
\equiv \left( \frac1{2M} \vee\etildei k_m(x)\right) \wedge \left( 1 - \frac1{2M} \right),
\end{equation}
\begin{equation}
\label{etilde}
\etildei k_m(x)
= \frac{S_m^{(k)}(x)}{N_k(x)},
\end{equation}
where $S_m^{(k)}(x) = \sum\limits_{i=1}^n w^{(k)}_i(X_i, x) \II(Y_i = m)$, $N_k(x) = 
\sum\limits_{i=1}^n w^{(k)}_i(X_i, x)$.
Then one can use the SSA procedure \cite{bs07} to construct aggregated estimates \( 
\that_1(x), \dots, \that_M(x) \).
The final prediction of the label at the point $x$ is given by the plug-in rule \eqref{plugin}.
The detailed description of the procedure for multiclass classification is given in 
Section \ref{numerical}.
We will refer to it as MSSA (short for Multiclass Spatial Stagewise Aggregation).

To show a consistency of the MSSA procedure, we will derive upper bounds for the 
generalization error $\p_{(X, Y)\sim\D}\left( Y \neq \widehat f(X) \big| S_n \right)$ of the 
classifier $\widehat f$, which hold in mean and with high probability over training 
samples $S_n$.
As a byproduct, we will provide convergence rates for the pointwise error 
$\max\limits_{1\leq m\leq M} |\that_m(x) - \tstar_m(x)|$ and obtain a user-friendly bound 
on the performance of the nearest neighbor estimates under mild assumptions.
Namely, along with \eqref{a1} and \eqref{a2}, we assume the following.
First, the functions $\eta_m(\cdot)$ are $(L, \alpha)$-Holder continuous, i. e. there exist 
$L > 0$ and $\alpha > 0$ such that for all $x, x' \in \x$ and $1 \leq m \leq M$ it holds
\begin{equation}
\label{a3}
\tag{A3}
|\eta_m(x) - \eta_m(x')| \leq L \| x - x' \|^\alpha .
\end{equation}
Second, since we deal with the problem of nonparametric classification, even the 
optimal rule can show poor performance in the case of a large dimension \( d \).
Low noise assumptions are usually used to speed up rates of convergence and allow 
plug-in classifiers to achieve fast rates.
We can rewrite
\begin{align}
\label{risk}
R(f)
& \notag
= 1 - \E_{(X, Y)\sim \D} \II(Y = f(X))
\\&
= 1 - \E_X \p(Y = f(X) | X)
= 1 - \E_X \eta_{f(X)}(X) .
\end{align}
In the case of binary classification, a misclassification often occurs, when \( \eta_1(X) 
\equiv \p( Y = 1 | X) \) is close to \( 1/2 \) with a high probability.
The well-known Mammen-Tsybakov noise condition \cite{mt99} ensures that such a 
situation appears rarely.
More precisely, it assumes that there exist non-negative constants \( B \) and \( \beta 
\) such that for all \( t > 0 \) it holds
\[
\label{mt}
\p_X \left( | 2\eta_1(X) - 1 | < t \right) \leq B t^{\beta} .
\]
This assumption can be extended to the multiclass case.
Let \( \eta_{(1)}(x) \geq \eta_{(2)}(x) \geq \dots \geq \eta_{(M)}(x) \) be the ordered values 
of \( \eta_1(x), \dots, \eta_M(x)\).
Then the condition \eqref{mt} for the multiclass classification can be formulated as 
follows (see \cite{agarwal, pr13}): there 
exist $B > 0$ and $\beta \geq 0$ such that for all $t > 0$ it holds
\begin{equation}
\label{a4}
\tag{A4}
\p_X \left( \eta_{(1)}(X) - \eta_{(2)}(X) < t \right) \leq Bt^\beta
\end{equation}
We will use this assumption to establish fast rates for the plug-in classifier \( \widehat 
f(X) \) in Section \ref{main_results}.

There are two more requirements we need: the minimal mass assumption and 
the tail assumption introduced in \cite{gkm16}.
The first one assumes that there exist $\varkappa > 0$ and $r_0 > 0$, such that for 
all $r \in (0, r_0]$ and $x \in \text{supp}(\p_X)$ it holds
\begin{equation}
\label{a5}
\tag{A5}
\p_X( X \in B(x, r) ) \geq \varkappa p(x) r^d,
\end{equation}
where $B(x, r)$ stands for the Euclidean ball of radius $r$ centered at $x$ and $p(x)$ is 
a density of the marginal distribution $\p_X$ with respect to the measure $\mu$.
The tail assumption admits that there are positive constants $C, \varepsilon_0$ and 
$p$ such that for every $\varepsilon \in (0, \varepsilon_0]$ it holds
\begin{equation}
\label{a6}
\tag{A6}
\p_X \left( p(X) < \varepsilon \right) \leq C\varepsilon^p .
\end{equation}
It was discussed in \cite{gkm16} (Theorem 4.1) that the conditions \eqref{a5} and 
\eqref{a6} are necessary for quantitative analysis of classifiers and cannot be removed.

One can pick out a simple case of a bounded away from zero density when for any $x 
\in\text{supp}(\p_X)$ it holds $p(x) \geq p_0 > 0$ with a positive constant $p_0$.
The most difficult points $x$ for classification with the nearest neighbor rule are those 
points, which are close to the decision boundary or where the density $p(x)$ approaches 
zero, because in this case a vicinity of $x$ may not contain the sample points at all.
One of the ways to control the misclassification error in the low-density region is to 
impose a modified smoothness condition on the regression function $\eta(\cdot)$, as it 
is done in \cite{cd14, dgw18}: they assume that there are constants $L > 0$ and 
$\alpha\in(0, 1]$, such that for all $x, x' \in \x$ it holds
\[
|\eta(x) - \eta(x')| \leq L \left( \p_X\big\{ B(x, \|x - x'\|) \big\} \right)^{\alpha/d} .
\]
This assumption ensures that in the regions with a small density $p(x)$ the function 
$\eta(x)$ is $(L', \alpha)$-Holder continuous with a small constant $L'$.
An approach, considered in \cite{gkm16}, uses assumptions \eqref{a5} and \eqref{a6} 
instead of the modified smoothness condition.
The assumption \eqref{a5} helps to control the minimal probability mass of the ball 
$B(x,r)$ in regions where the density $p(x)$ is close to zero.
A curious reader can ensure that all the results we formulate will also hold if $p(x)$ and 
$\varkappa$ in \eqref{a5} are replaced with $p_0$ and $\mu(B(0,1))$ respectively in the 
case of a bounded away from zero density $p(x)$.
Also, note that in this case, the assumption \eqref{a6} is satisfied with $\varepsilon_0 
< \min\{1, p_0\}$ and the power $p = \infty$.

We proceed with several examples of distributions when the tail assumption \eqref{a6} 
holds.
For instance, univariate Gaussian $\mathcal N(\mu, \sigma^2)$, exponential distribution 
$\text{Exp}(\lambda)$, gamma-distribution $\text{Gamma}(k, \lambda)$, Cauchy and 
Pareto $\text{P}(k, 1)$ distributions meet $\eqref{a6}$ with the powers $1$, $1$, 
$1+\varepsilon$ (with arbitrary $\varepsilon > 0$), $1/2$ and $k/(k+1)$ respectively (see 
\cite{gkm16}, Example 4.1 for the details).
A special case, in which one may be interested in, is the case when $\text{\rm 
	supp}(\p_X)$ is compact.
In this case,
\[
\p_X\left( p(X) < \varepsilon \right)
= \int\limits_{\text{\rm supp}(\p_X)} \II \left( p(X) < \varepsilon \right) p(x) dx
\leq \varepsilon \int\limits_{\text{\rm supp}(\p_X)} dx
= \varepsilon \mu\left( \text{\rm supp}(\p_X) \right),
\]
so \eqref{a6} is satisfied with $p=1$ and $C = \mu\left( \text{\rm supp}(\p_X) \right)$, 
where $\mu$ stands for the Lebesgue measure.
In general, the assumption \eqref{a6} admits that $\p_X$ has an unbounded support.
For this case, we give a simple sufficient condition to check \eqref{a6}.

\begin{prpstn}
	\label{prop1}
	Let $X \in \R^d$ be such that $\E\|X\|^r < \infty$.
	Then $X$ satisfies \eqref{a6} with $p = r/(r+d)$ and
	\[
	C =  \left( \left(\frac rd \right)^{\frac d{r+d}} + \left(\frac dr \right)^{\frac r{r+d}} 
	\right) \omega_d^{\frac r{r+d}} \left(\E\|X\|^r\right)^{\frac 
		d{r+d}},
	\]
	where $\omega_d$ stand for the Lebesgue measure of the unit ball in $\R^d$.
\end{prpstn}

\begin{proof}
	The proof of the proposition is straightforward:
	\begin{align*}
	&
	\p_X\left( p(X) < \varepsilon \right)
	= \int\limits_{\R^d} \II\left( p(X) < \varepsilon \right) p(x) dx
	\\&
	= \int\limits_{x\in B(0,R)} \II\left( p(X) < \varepsilon \right) p(x) dx
	+ \int\limits_{x\notin B(0,R)} \II\left( p(X) < \varepsilon \right) p(x) dx 
	\\&
	\leq \varepsilon R^d \omega_d + \int\limits_{x\notin B(0,R)} \frac{\|x\|^r}{R^r} p(x) 
	dx
	\leq \varepsilon R^d \omega_d + \frac{\E\|X\|^r}{R^r} .
	\end{align*}
	Taking $R^{r+d} = r\E\|X\|^r / (d \varepsilon \omega_d)$ to minimize the expression 
	in the right hand side, we obtain
	\[
	\p_X\left( p(X) < \varepsilon \right)
	\leq 
	\left( \left(\frac rd \right)^{\frac d{r+d}} + \left(\frac dr \right)^{\frac r{r+d}} 
	\right) \left( \omega_d \varepsilon \right)^{\frac r{r+d}} \left(\E\|X\|^r\right)^{\frac 
		d{r+d}} .
	\]
\end{proof}
In what is going further, we require $p$ in \eqref{a6} to be larger than 
$\alpha/(2\alpha+d)$.
By Proposition \ref{prop1}, any $\p_X$, such that $\E\|X\|^r < \infty$ for some $r > 
\alpha d/(\alpha + d)$, satisfies \eqref{a6} with $p > \alpha/(2\alpha+d)$.

\section{The algorithm}
\label{algorithm}

In this section, we present the multiclass spatial stagewise aggregation (MSSA) 
procedure, which is precisely formulated in Algorithm \ref{SSA}.
The procedure takes a sequence of integers $\{ n_k : 1 \leq k \leq K\}$, which fulfills 
\eqref{a2}, a training sample $S_n = \{ (X_i, Y_i) : 1 \leq i \leq n \}$, a test point $x \in 
\x$ and a set of positive numbers $\left\{ z_k : 1 \leq k \leq K \right\}$.
The numbers $z_1, \dots, z_K$ will be referred to as critical values.
This name is not occasional since the original spatial stagewise aggregation procedure is 
tightly related to hypothesis testing.
More details can be found in \cite{bs07}.
It is important to mention that performance of the MSSA procedure crucially depends on 
a choice of the critical values $z_k$, $1 \leq k \leq K$.
At first glance, one can think that the problem of tuning of so large number of 
parameters is very time consuming and impracticable.
However, in Section \ref{numerical} with numerical experiments we provide a simple 
tuning procedure leading to a proper choice of the critical values.

\begin{lgrthm}
	Multiclass Spatial Stagewise Aggregation (MSSA)
	\label{SSA}
	\begin{algorithmic}[1]
		\State
		Given a sequence of integers \( \{ n_k : 1 \leq k \leq K\}\) fulfilling \eqref{a2}, a set 
		of critical values\( \left\{ z_k : 1 \leq k \leq K \right\} \) , a training sample $S_n = \{ 
		(X_i, Y_i) : 1 \leq i \leq n \}$ and a test point $x \in \x$, do the following:
		\For{ \( m \textbf{ from } 1 \textbf{ to } M \) }
		\State For each $k$ from $1$ to $K$ compute the weights $w^{(k)}_i = 
		w^{(k)}_i(X_i, x)$, $1 \leq i \leq n$, according to the formula \eqref{weights2} 
		\\ \hspace{1em} 
		with a kernel $\mathscr K$ satisfying \eqref{a1} and calculate $\ttildei k_m(x)$ 
		according to \eqref{ttilde} and \eqref{etilde}.
		\State Put $\thati 1_m(x) = \ttildei 1_m(x)$.
		\For{\( k \textbf{ from } 2 \textbf{ to }  K \)}
		\State Compute $N_k(x) = \sum\limits_{i=1}^n w^{(k)}_i(X_i, x)$ and
		\\ \hspace{2.7em}
		$\kl{\ttildei k_m(x)}{\thati{k-1}_m(x)} = \ttildei k_m(x) 
		\log\frac{\ttildei k_m(x)}{\thati{k-1}_m(x)} + \left(1 - \ttildei k_m(x) \right) 
		\log\frac{1 - \ttildei k_m(x)}{1 - \thati{k-1}_m(x)}$.
		\State Find $\gamma_k = \II \left( N_k(x) \kl{\ttildei k_m(x)}{\thati{k-1}_m(x)} \leq 
		z_k \right)$.
		\State Update the estimate $\thati k_m(x) = \gamma_k \ttildei k_m(x) + (1 - 
		\gamma_k) \thati{k-1}_m(x)$.
		\EndFor
		Put the final estimate \( \that_m(x) = \thati K_m(x) \).
		\EndFor
		\State\Return the predicted label \( \widehat f(x) = \argmax\limits_{1 \leq m \leq 
			m}\left\{\that_m(x)\right\} \).
	\end{algorithmic}
\end{lgrthm}

We also emphasize that, by construction, $\ttildei k_m(x) \in \left[ 1/(2M), 1 - 1/(2M) 
\right]$ and, therefore, $\thati k_m(x)$ also belongs to $\left[ 1/(2M), 1 - 1/(2M) \right]$ 
and $\kl{\ttildei k_m(x)}{\thati{k-1}_m(x)}$ is defined correctly.
In fact, $\kl{\ttildei k_m(x)}{\thati{k-1}_m(x)}$ is nothing but the Kullback-Leibler 
divergence between two Bernoulli distributions with parameters $\ttildei k_m(x)$ and 
$\thati{k-1}_m(x)$ respectively.

Concerning the computational time of the MSSA procedure, the assumption \eqref{a2} 
ensures that $K = O(\log n)$ and then it requires $O \left( Mn\log n \right)$ operations 
to compute nearest neighbor estimates for all classes and \( O( \log n) \) operations to 
aggregate them.
As a result, the computational time of the procedure, consumed for a prediction of the 
label of one test point, is $O \left( Mn\log n \right)$.
If there are several test points, then the computations can be done in parallel.

\section{Theoretical properties of the MSSA procedure}
\label{main_results}

\subsection{Main result}

\begin{thrm}
	\label{main}
	Let the conditions \eqref{a1} -- \eqref{a5} hold and let \eqref{a6} hold with $p > 
	\alpha/(2\alpha + d)$. 
	Choose the parameters $z_1, \dots, z_K$ according to the formula
	\begin{equation}
	\label{z_ks}
	z_k = \frac{8M^2}{u_0} \log \frac{12KM}{\delta_*}, \quad 1 \leq k \leq K,	
	\end{equation}
	where
	\begin{equation}
	\label{delta}
	\delta_* =
	\begin{cases}
	\left( \frac{M^3\log n}{np_0} \right)^{\frac{\alpha(2+\beta)}{2\alpha + d}}, 
	\quad \text{if } \exists \, p_0 : p(x) \geq p_0 \, \forall \, x \in 
	\text{supp}(\p_X),\\
	\psi_*^{r_*}, \quad \text{otherwise},
	\end{cases}
	\end{equation}
	with $r_* = \log \psi_*^{-1}$ and
	\[
	\psi_* = \left( \frac{M^3 \log^2 n}n \right)^{\frac{\alpha}{\alpha\beta/p + 2\alpha + 
			d}} .
	\]
	Then, if the sample size $n$ is sufficiently large, for the MSSA estimates 
	$\that_1(\cdot), \dots, \that_M(\cdot)$, the excess risk of the plug-in classifier 
	$\widehat f(X) = \argmax\limits_{1 \leq m \leq M} \that_m(X)$ is bounded by
	\begin{equation}
	\label{main1}
	\E_{S_n} \mathcal E(\widehat f) \lesssim
	\begin{cases}
	\left( \frac{M^3\log n}{np_0} \right)^{\frac{\alpha(1+\beta)}{2\alpha + d}}, 
	\quad \text{if } \exists \, p_0 : p(x) \geq p_0 \, \forall \, x \in 
	\text{supp}(\p_X),\\
	\left( \frac{M^3\log^2 n}n \right)^{\frac{\alpha(1+\beta)}{\alpha\beta/p + 
			2\alpha + d}}, \quad \text{otherwise} .
	\end{cases}
	\end{equation}
	Moreover, for any $\delta \in (0, 1)$, if
	\[
	z_k = \frac{8M^2}{u_0} \log \frac{12KM}{\delta}, \quad 1 \leq k \leq K,
	\]
	on an event with probability at least $(1 - \delta)$ over training samples, it holds
	\begin{equation}
	\label{main3}
	\mathcal E(\widehat f)
	\leq \p_X(\widehat f(X) \neq f^*(X))
	\lesssim \delta + \left( \frac{M^3 \log(12KM/\delta)}n 
	\right)^{\frac{\alpha\beta}{\alpha\beta/p + (2\alpha + d)}}.
	\end{equation}
\end{thrm}
Here and further in the paper the relation $g(n) \lesssim h(n)$ means that there exists a 
universal constant $c > 0$ such that $g(n) \leq c h(n)$ for all $n \in \mathbb N$.

There are some comments we have.
First, the rates \eqref{main1} are optimal up to a logarithmic factor (see \cite{at07}, 
Theorem 3.2 for the case of bounded away from zero density, \cite{at07}, Theorem 4.1 
for the case of bounded support (i.e. $p=1$ in \eqref{a6}), and \cite{gkm16}, Theorem 
4.5 for the general case).
Second, in the case of a bounded away from zero density one can take $p = \infty$.
Then the inequality \eqref{main3} transforms into
\[
\p_X(\widehat f(X) \neq f^*(X)) \lesssim \delta + \left( 
\frac{M^3 \log(12KM/\delta)}n \right)^{\frac{\alpha\beta}{2\alpha + d}},
\]
which revisits the result  of Theorem 7 in \cite{cd14}.

\subsection{Comparison with the nearest neighbor rule}

\begin{thrm}
	\label{knnbound}
	Assume \eqref{a1}, \eqref{a3} and \eqref{a5}.
	Fix any $m$, $1 \leq m \leq M$, and a test point $x \in \x$.
	Then, for the weighted nearest neighbor estimate $\etilde^w_m(x)$ defined by 
	\eqref{wknn} and \eqref{weights}, with probability at least $(1 - 
	\delta)$ over all training samples, it holds
	\[
	|\eta_m(x) - \etilde^w_m(x)|
	\leq \frac L{\left(n\varkappa p(x)\right)^{\alpha/d}} \big( 2\mathsf k + 
	4\log(2/\delta) \big)^{\alpha/d} + \sqrt{ \frac{\log(4/\delta)}{\mathsf k} },
	\]
	for any $\mathsf k$ and $\delta \in (0, 1)$, satisfying
	\[
	\left( \frac{2\mathsf k + 4\log(1/\delta)}{n\varkappa p(x)} \right)^{\alpha/d} \leq 
	r_0 .
	\]
\end{thrm}
The proof of this result is moved to Appendix \ref{knnboundproof}.
The bound in Theorem \ref{knnbound} improves the result for the nearest neighbor 
regression obtained in \cite{dgw18} since it controls large deviations of $|\eta_m(x) - 
\etilde^w_m(x)|$ rather than its mean value.
For the case of a bounded away from zero density, Theorem \ref{knnbound}  and the 
union bound immediately yield
\[
\E_{S_n} \E_X \max\limits_{1\leq m\leq M}|\eta_m(X) - \etilde^w_m(X)|^r \lesssim 
\left( \frac{\mathsf k \log M}n \right)^{\alpha r/d} + \left( \frac{\log M}{\mathsf k} 
\right)^{r/2}
\]	
for any $r > 0$.
This, together with Lemma \ref{fast_rate} below, implies a bound for the 
$\mathsf k$-nearest neighbors classifier $\widehat f^{(\mathsf k-NN)}(x) = 
\argmax\limits_{1\leq m\leq M} \etilde^w_m(x)$:
\[
\E_{S_n} \mathcal E\left( \widehat f^{(\mathsf k-NN)}(x) \right) \lesssim \left( 
\frac{\log M}n \right)^{\frac{\alpha(1+\beta)}{2\alpha + d}} ,
\]
provided that $\mathsf k \asymp n^{2\alpha/(2\alpha + d)}$.

In the case of the bounded away from zero density, the nearest neighbor rule attains the 
minimax rate $n^{-(1 + \beta)/(2\alpha + d)}$, while the MSSA classifier has an 
additional logarithmic factor.
It can be easily explained by the fact that in the case $p(x) \geq p_0$, it is enough to 
take only one number of neighbors $\mathsf k \asymp n^{d/(2\alpha + d)}$ for all points 
$x \in \x$.
At the same time, the MSSA procedure aggregates several nearest neighbor estimates 
and the factor $\log n$ can be considered as a payment for adaptation.
Nevertheless, MSSA is capable to adapt to an unknown smoothness parameter 
$\alpha\in(0,1]$ from the condition \eqref{a3}, while the optimal choice of the smoothing 
parameter $\mathsf k$ of the classifier $\widehat f^{(\mathsf k-NN)}$ is based on the 
knowledge of $\alpha$.

The situation is completely different in the case of a general density, fulfilling \eqref{a5} 
and \eqref{a6}.
In \cite{gkm16} (Theorems 4.3 and 4.5), it was shown that a universal choice of $\mathsf 
k$ for all points $x\in \x$ leads to a suboptimal rate $n^{-\frac{\alpha(1+\beta)}{\alpha(1 
		+ \beta)/p + 2\alpha + d}}$, while Theorem \ref{main} guarantees that the MSSA 
classifier has a minimax rate of convergence up to a logarithmic factor.
It was also shown in \cite{gkm16} (Theorems 4.4 and 4.5) that a point-dependent choice 
$\mathsf k(x) \asymp (n p(x))^{2\alpha/(2\alpha + d)}$ leads to the same rate $\big( (\log 
n)/n \big)^{\frac{\alpha(1+\beta)}{\alpha\beta/p + 2\alpha + d}}$, as for the MSSA 
classifier up to a logarithmic factor.
However, it is not clear how to implement such a choice of $\mathsf k$ in practice, since 
a prior knowledge of the density $p(x)$ is required.
Of course, one can try to estimate $p(x)$ but the density estimates are susceptible to 
the curse of dimensionality.
In our turn, in Section \ref{numerical}, we describe a simple procedure of tuning 
parameters of MSSA.
Moreover, by Theorem 3.1, the choice of critical values is the same for all test points, 
while the estimate of $p(x)$ must be recomputed at each test point $x$.

\section{Numerical experiments}
\label{numerical}
This section serves to illustrate the numerical performance of the proposed MSSA 
procedure on the artificial and real datasets.
First we specify the choice of tuning parameters, then present the results.

\subsection{Parameter tuning by propagation}
Performance of the procedure critically depends on the choice of parameters
\( z_{k} \). 
We apply here the propagation approach originating from \cite{SV2009}.
The basic idea is to ensure the desired properties of the method 
in a specific homogeneous situation. 
Let \( x \in \x \) be a fixed test point.
We generate artificial labels $\widetilde Y_1, \dots, \widetilde Y_n$, which are sampled 
independently according to the distribution $\text{Bernoulli}\left(\frac12\right)$.
In this case, $\eta_1(x) = \p(Y = 1 | X = x) \equiv 1/2$.
Now, the proof of Lemma \ref{mssabound} gives an insight, how to choose the critical 
values $z_k$: in the homogeneous situation $\eta_1(x) \equiv 1/2$, an event $\{ \exists \, k 
: \thati k_1(x) \neq \ttildei k_1(x) \}$ has to occur with a small probability.
Such property of the MSSA procedure is called propagation.
The preliminary critical values \( \widetilde z_{2},\ldots, \widetilde z_{K} \) are computed 
sequentially. 
Suppose that \( \widetilde z_{2},\ldots,\widetilde z_{k-1} \) have been already fixed for \( k 
\geq 1 \). 
This allows to compute \( \thati{k-1}_1(x) \) and the test statistic \( T_k = N_k(x) \kl{\ttildei 
	k_1(x)}{\thati{k-1}_1(x)} \). 
Now \( \widetilde z_{k} \) is defined from the condition
\[
\p\bigl( T_{k} > \widetilde z_{k} \cond T_{2} \leq \widetilde z_{2},\ldots T_{k-1} \leq 
\widetilde z_{k-1} \bigr)
\leq \delta/K .
\]
for some small $\delta \in (0,1)$, e.g. \( \delta = 0.1 \). 
This condition is checked by the Monte-Carlo simulations 
for the artificial dataset $\widetilde S_n 
= \{ (X_i, \widetilde Y_i) : 1 \leq i \leq n \}$.
After that we choose the critical values $z_1, \dots, z_K$ in the form $z_k = c \widetilde 
z_k$ for all $k$ from $1$ to $K$.
The constant $c$ is chosen by cross validation.
The Monte-Carlo simulations are performed only for one test point, because, due to 
Theorem \ref{main}, the choice of $z_k$'s is universal for all test points.

\subsection{Experiments on artificial datasets}

\begin{figure}[H]
	\begin{minipage}{0.32\linewidth}
		\noindent
		\centering
		\includegraphics[width=\linewidth]{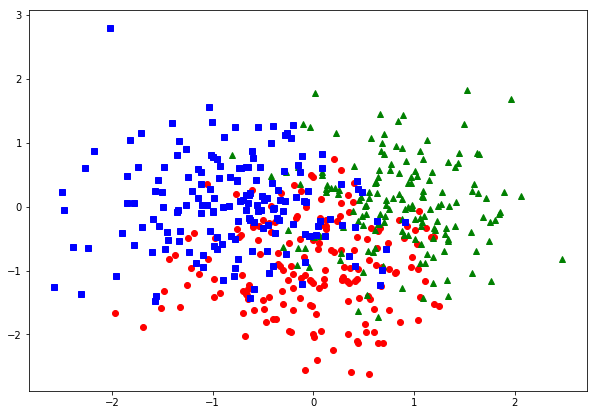}
	\end{minipage}
	\hfill
	\begin{minipage}{0.32\linewidth}
		\noindent
		\centering
		\includegraphics[width=\linewidth]{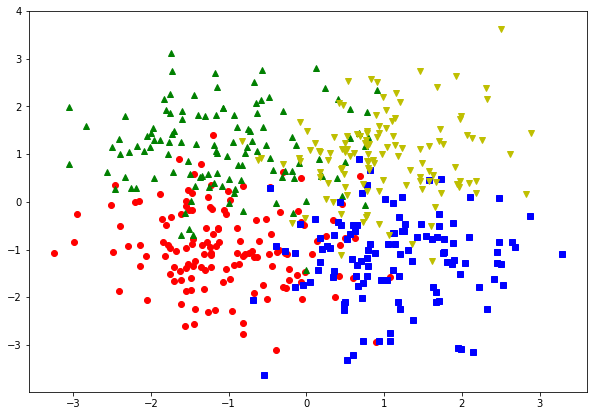}
	\end{minipage}
	\hfill
	\begin{minipage}{0.32\linewidth}
		\noindent
		\centering
		\includegraphics[width=\linewidth]{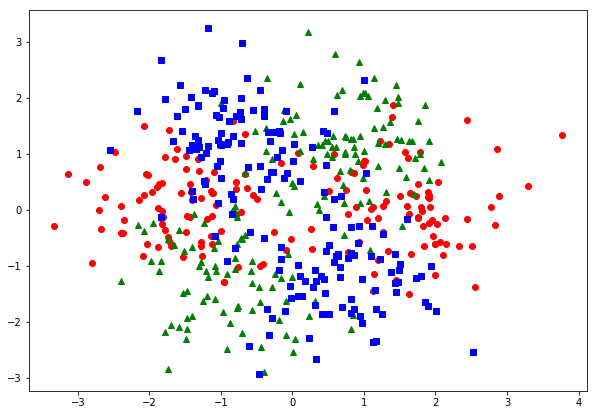}
	\end{minipage}
	\caption{Sample realizations in the first (left, \( M = 3 \) classes, \( n=500 \) points), 
		the second (center, \( M = 4 \) classes, \( n=500 \) points) and the third (right, \( 
		M = 3 \) classes, \( n=500 \) points) experiments with artificial datasets.}
	\label{sample}
\end{figure}

\begin{figure}[H]
	\begin{minipage}{0.32\linewidth}
		\noindent
		\centering
		\includegraphics[width=\linewidth]{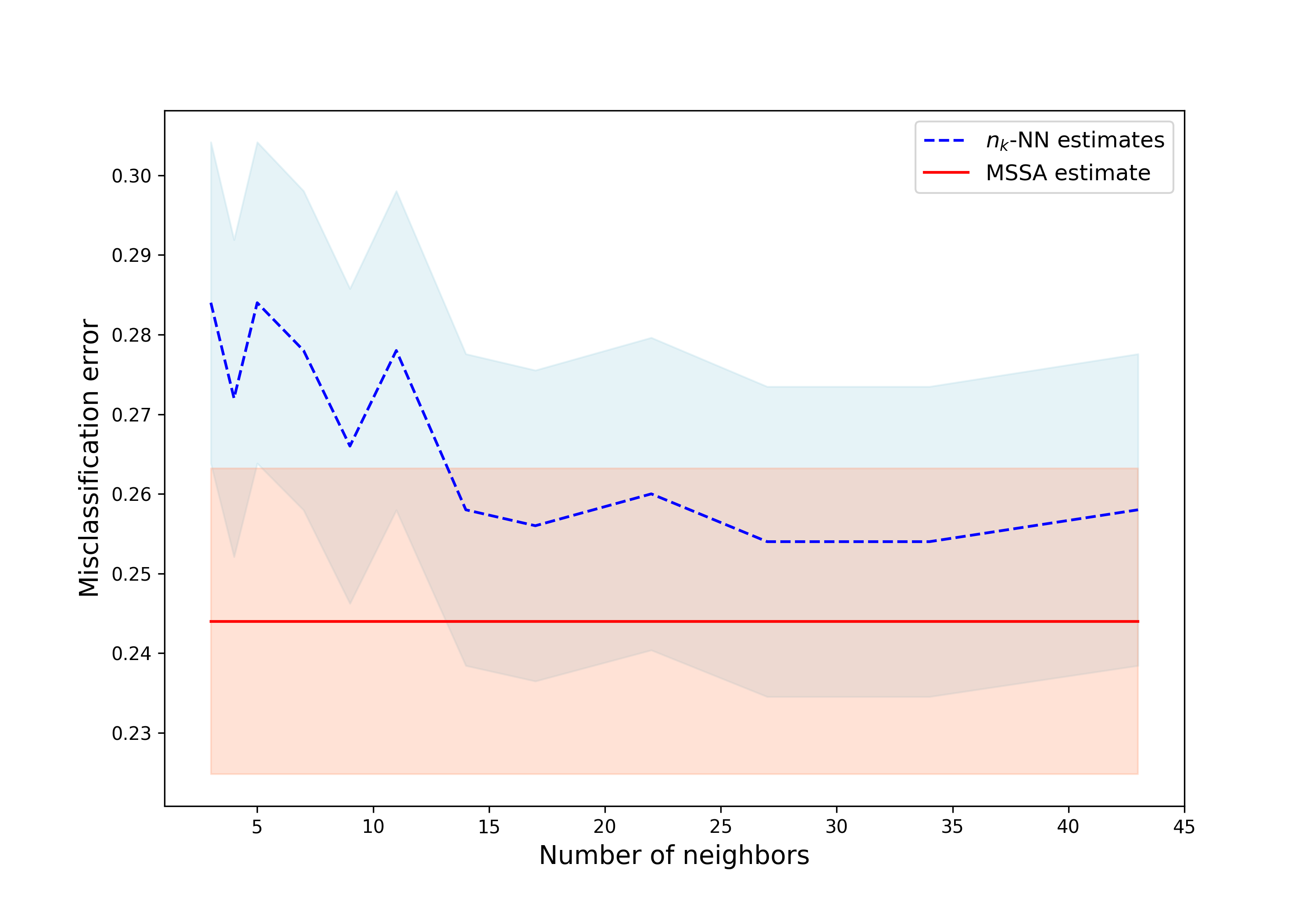}
	\end{minipage}
	\hfill
	\begin{minipage}{0.32\linewidth}
		\noindent
		\centering
		\includegraphics[width=\linewidth]{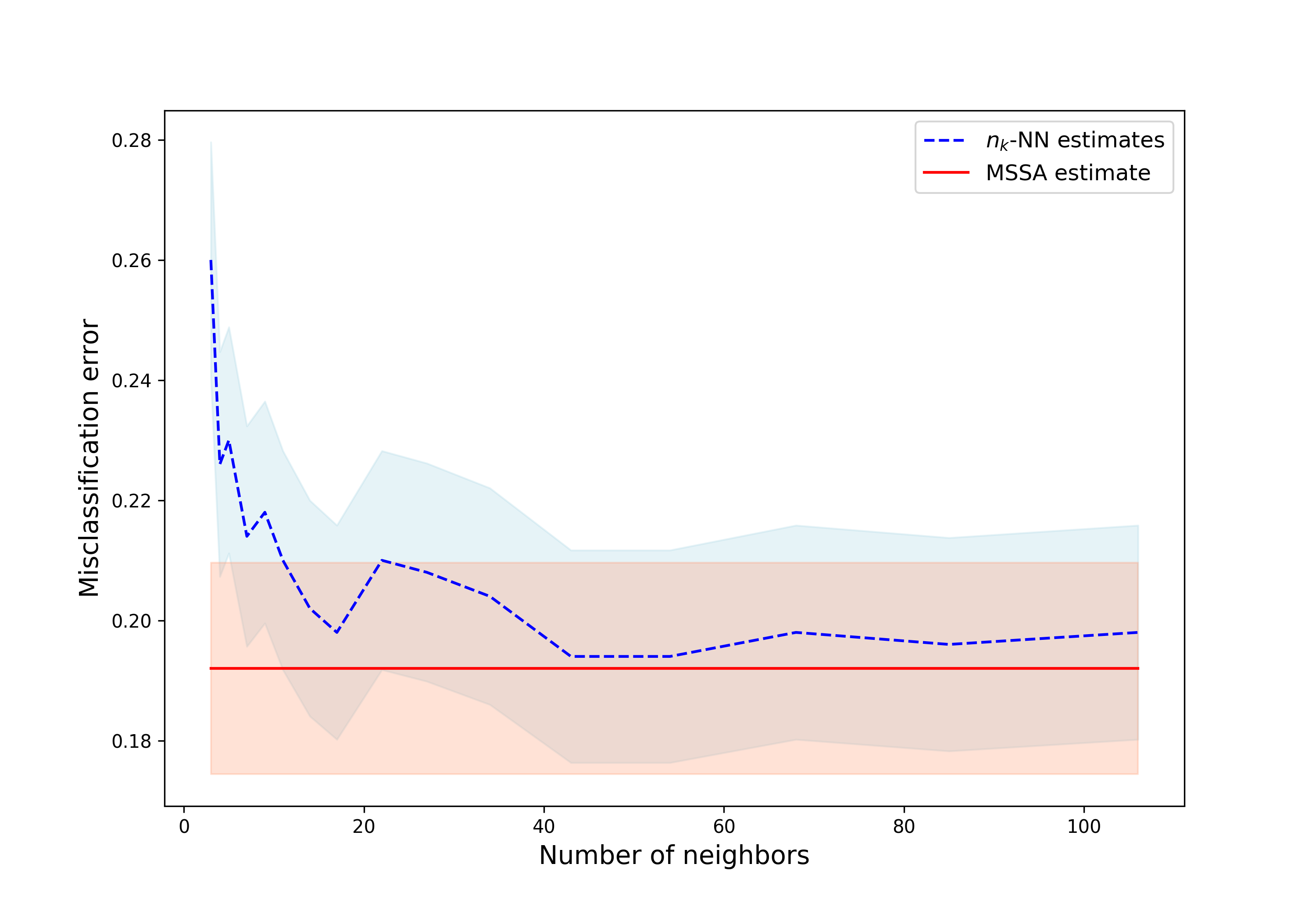}
	\end{minipage}
	\hfill
	\begin{minipage}{0.32\linewidth}
		\noindent
		\centering
		\includegraphics[width=\linewidth]{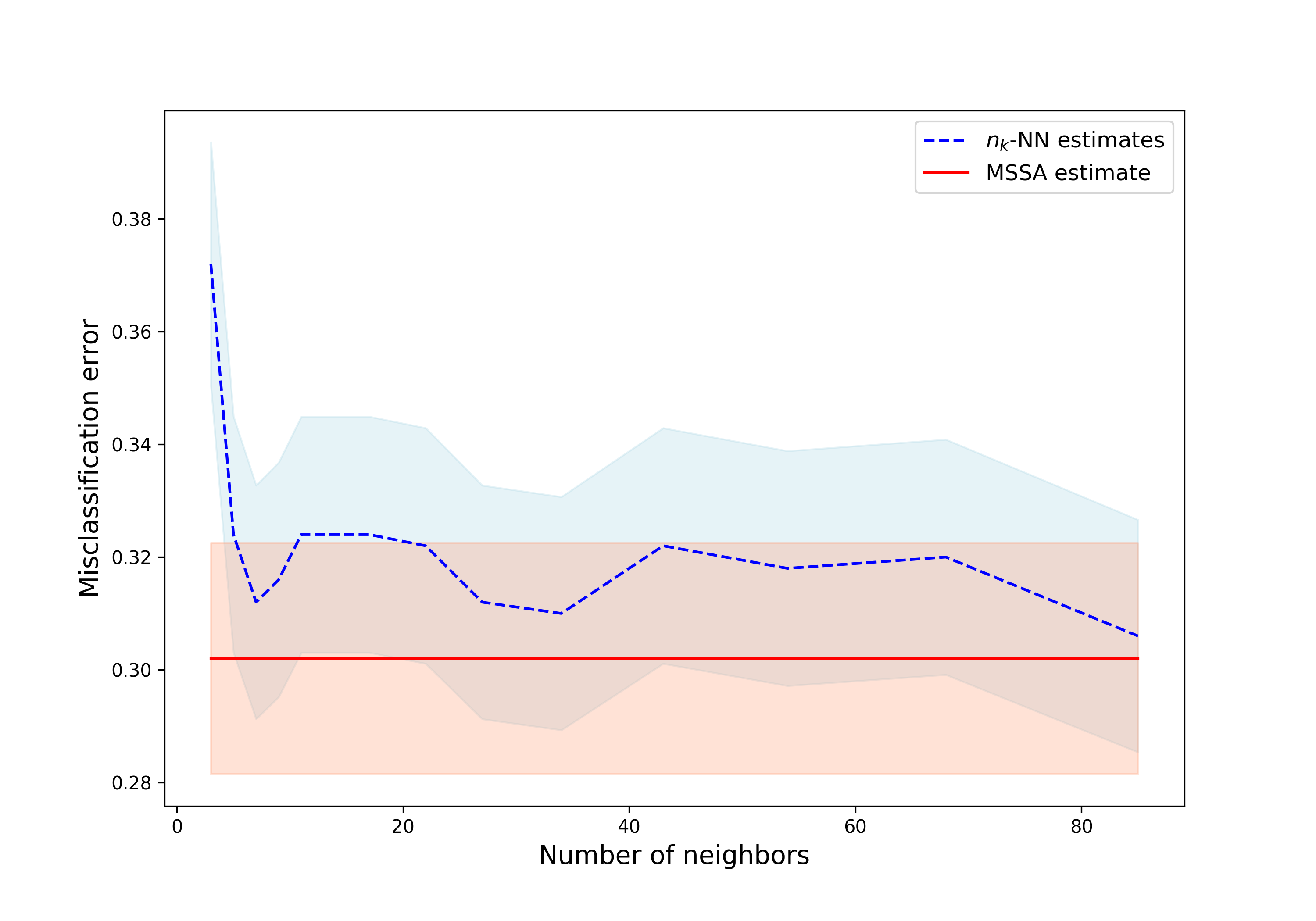}
	\end{minipage}
	\caption{Leave-one-out cross-validation errors for the weighted nearest-neighbor 
		classifiers (the dashed line) in the first (left), the second (center) and the third 
		(right) experiments. The solid line corresponds to the LOO CV error of the MSSA 
		classifier. The shaded regions reflect standard deviations.}
	\label{res}
\end{figure}

We start with presenting the performance of MSSA on artificial datasets.
We generate points from a mixture model: $p(x | Y = m) = p_m(x)$, $\p(Y = m) = \pi_m$.
Then the density of \( X \) is given by $p(x) = \sum\limits_{m=1}^M \pi_m p_m(x)$,
and the Bayes rule is defined as $f^*(X) = \argmax\limits_{1\leq m\leq M} \pi_m p_m(x)$.
We provide results for three different experiments.
The information about them is summarized in Table \ref{artificial} and sample 
realizations are displayed in Figure \ref{sample}.
\begin{table}[ht]
	\noindent
	\begin{adjustwidth}{-0.5cm}{}
		\begin{tabular}{|p{7em}|p{14em}|p{14em}|p{17em}|}
			\hline
			& Experiment 1 & Experiment 2 & Experiment 3 \\
			\hline
			Sample size, $n$ & 500 & 500 & 500 \\
			\hline
			Number of classes, $M$ &  3 & 4 & 3 \\
			\hline
			Prior class probabilities, $\pi_m$& 1/3, 1/3, 1/3 & 1/4, 1/4, 1/4, 1/4 
			& 1/3, 1/3, 1/3 \\
			\hline
			Class densities, $p_m(x)$ & $p_1(x) = \phi\left(x, [0, -1], 0.5 
			I_2\right)$, $p_2(x) = \phi\left(x, [\sqrt3/2, 0], 0.5 I_2\right)$, $p_3(x) = 
			\phi\left(x, [-\sqrt3/2, 0], 0.5 I_2\right)$ & $p_1(x) = \phi\left(x, [-1, -1], 0.7 
			I_2\right)$, $p_2(x) = \phi\left(x, [1, -1], 0.7 I_2\right)$, $p_3(x) = \phi\left(x, 
			[-1, 1], 0.7 I_2\right)$, $p_4(x) = \phi\left(x, [1, 1], 0.7 I_2\right)$ & $p_1(x) = 
			0.5\phi\left(x, [-1, 0], 0.5 I_2\right) \newline + 0.5\phi\left(x, [1, 0], 0.5 
			I_2\right)$,\newline $p_2(x) = 0.5\phi\left(x, [0.5, \sqrt3/2], 0.5 I_2\right) 
			\newline + 0.5\phi\left(x, [-0.5, -\sqrt3/2], 0.5 I_2\right)$, \newline$p_3(x) = 
			0.5\phi\left(x, [-0.5, \sqrt3/2], 0.5 I_2\right) \newline + 0.5\phi\left(x, [0.5, 
			-\sqrt3/2], 0.5 I_2\right)$ \\
			\hline
			Number of neighbors, $n_k$ & $n_k = \lfloor 3 \cdot 1.25^k 
			\rfloor$,  
			$0 \leq k \leq 11$ & $n_k = \lfloor 3 \cdot 1.25^k\rfloor$, $0 \leq k 
			\leq 15$ & $n_k = \lfloor 3 \cdot 1.25^k\rfloor$, $0 \leq k \leq 14$ \\
			\hline
			Localization kernel, $\mathscr K(t)$ & rectangular, \newline $\mathscr K(t) 
			= \II(0 \leq t \leq 1)$& rectangular, \newline $\mathscr K(t) = \II(0 \leq t \leq 
			1)$ & rectangular, \newline $\mathscr K(t) = \II(0 \leq t \leq 1)$ \\
			\hline
		\end{tabular}
	\end{adjustwidth}
	\caption{Information about artificial datasets. $\phi(\cdot, \mu, \Sigma)$ stands for 
		the density of the Gaussian distribution $\mathcal N(\mu, \Sigma)$.}
	\label{artificial}
\end{table}
For example, in the first experiment, the sample consists of $n = 500$ points, each of 
them belongs to one of $M = 3$ classes, and the prior class probabilities $\pi_m$, $1 
\leq m \leq 3$, are equal to $1/3$.
Class densities $p_1(x)$, $p_2(x)$ and $p_3(x)$ were taken $\phi\left(x, [0, -1], 0.5 
I_2\right)$, $\phi\left(x, [\sqrt3/2, 0], 0.5 I_2\right)$ and $\phi\left(x, [-\sqrt3/2, 0], 0.5 
I_2\right)$ respectively, where \( \phi(x, \mu, \Sigma) \) stands for the density of a 
Gaussian random vector with the mean \( \mu \) and the variance \( \Sigma \).
Next, we took the sequence of integers $n_k = \lfloor 3 \cdot 1.25^k \rfloor$, $0 \leq k 
\leq 11$, and considered \( n_k \)-nearest-neighbors estimates with the rectangular 
kernel $\mathscr K(t) = \II(0 \leq t \leq 1)$.
We computed leave-one-out cross-validation errors for the MSSA classifier and all 
$n_k$-nearest neighbors classifiers.
The second and the third experiments on artificial datasets were carried out in the same 
way.
The results, which are shown on Figure \ref{res}, indicate that even the best 
$n_k$-nearest neighbors classifier is outperformed by the properly tuned MSSA 
classifier.

\subsection{Experiments on the real datasets}

We proceed with experiments on datasets from the UCI repository \cite{uci}: Ecoli, Iris, 
Glass, Pendigits, Satimage, Seeds, Wine and Yeast.
Short information about these datasets is given in Table \ref{datasets}.

\begin{table}[H]
	\begin{center}
		\begin{tabular}{|c|c|c|c|c|p{15em}|}
			\hline
			Dataset & Train & Test & Attributes & Classes & Class distribution 
			(in \%)\\
			\hline
			Ecoli & 336 & -- & 7 & 8 & 42.6, 22.9, 15.5, 10.4, 5.9, 1.5, 0.6, 0.6\\
			\hline
			Iris & 150 & -- & 4 & 3 & 33.3, 33.3, 33.3\\
			\hline
			Glass & 214 & -- & 9 & 6 & 32.7, 35.5, 7.9, 6.1, 4.2, 13.6\\
			\hline
			Pendigits & 7494 & 3498 & 16 & 10 & 10.4, 10.4, 10.4, 9.6, 10.4, 9.6, 9.6, 
			9.6, 10.4, 9.6, 9.6 \\
			\hline
			Satimage & 4435 & 2000 & 36 & 6 & 24.1, 11.1, 20.3, 9.7, 11.1, 23.7 \\
			\hline
			Seeds & 210 & -- & 7 & 3 & 33.3, 33.3, 33.3 \\
			\hline
			Wine & 178 & -- & 13 & 3 & 33.1, 39.8, 26.9 \\
			\hline
			Yeast & 1484 & -- & 8 & 10 & 16.4, 28.1, 31.2, 2.9, 2.3, 3.4, 10.1, 2.0, 1.3, 0.3 
			\\
			\hline
		\end{tabular}
	\end{center}
	\caption{Information about datasets from the UCI repository \cite{uci}.}
	\label{datasets}
\end{table}

We compare the performance of the MSSA algorithm with the oracle choice of the 
nearest neighbor estimate.
For Pendigits and Satimage datasets, we calculated misclassification error on the 
test dataset, for all other datasets we used leave-one-out cross-validation.
Results of our experiments are shown in Table \ref{results}, best ones are boldfaced.
From Table \ref{results}, one can observe that MSSA works as good as $\mathscr k$-NN 
rule with the best choice of nearest neigbors and even slightly outperforms ordinary 
nearest neighbor rule in most situations.

\begin{table}[H]
	\begin{center}
		\begin{tabular}{|c|c|c|}
			\hline
			Dataset & MSSA & Best nearest neighbor classifier\\
			\hline
			Ecoli & $\bf 12.8 \pm 1.8$ & $13.4 \pm 1.9$\\
			\hline
			Iris & $0.0 \pm 0.0$ & $0.0 \pm 0.0$ \\
			\hline
			Glass & $\bf 27.6 \pm 3.0$ & $28.0 \pm 3.0$ \\
			\hline
			Pendigits & $2.2 \pm 0.2$ & $2.2 \pm 0.2$ \\
			\hline
			Satimage & $\bf 9.3 \pm 0.6$ & $9.6 \pm 0.7$ \\
			\hline
			Seeds & $6.7 \pm 1.7$ & $6.7 \pm 1.7$ \\
			\hline
			Wine & $2.2 \pm 1.1$ & $2.2 \pm 1.1$ \\
			\hline
			Yeast & $\bf 39.6 \pm 1.3$ & $39.8 \pm 1.3$ \\
			\hline
		\end{tabular}
	\end{center}
	\caption{Leave-one-out cross-validation errors (in \%) with standard deviations for 
		datasets from the UCI repository. The best results are boldfaced.}
	\label{results}
\end{table}

\bibliographystyle{abbrv}
\bibliography{references}

\appendix

\section{Proof of Theorem \ref{main}}
\label{proof}

The proof of Theorem \ref{main} is divided into several steps.
On the first one, we discuss nice properties of the MSSA 
estimates $\that_m(x)$, $1 \leq m\leq M$.
Next, we focus on the MSSA plug-in classifier $\widehat f(x) = \argmax\limits_{1\leq 
	m\leq M} \that_m(x)$.
In Section \ref{step2}, we study the case of a bounded away from zero density and prove 
the first part of the upper bound \eqref{main1} for the mean excess risk $\E_{S_n} 
\mathcal E(\widehat f)$. 
Then, in Section \ref{step3}, we extend our analysis to the case of a general density 
$p(x)$, which fulfils the minimal mass assumption \eqref{a5} and the tail assumption 
\eqref{a6}.
Finally, in Section \ref{step4}, we obtain the bound \eqref{main3} on the excess risk 
$\mathcal E(\widehat f)$, which holds on an event with high probability.

\subsection{Step 1: pointwise guarantees for MSSA estimates}
\label{step1}

Theorem \ref{knnbound}, the union bound and 1-Lipschitzness of the function 
$\varphi(\cdot)$ immediately yield
\begin{crllr}
	\label{knn_corollary}
	Under assumptions of Theorem \ref{knnbound}, we have
	\begin{align*}
		|\theta_m(x) - \ttildei k_m(x)|
		&
		\leq \frac L{\left(n\varkappa p(x)\right)^{\alpha/d}} \big( 2n_k + 4\log(6KM/\delta) 
		\big)^{\alpha/d}
		\\&
		+ \sqrt{ \frac{\log(12KM/\delta)}{n_k} }
	\end{align*}
	simultaneously for all $1 \leq m \leq M$ and $1 \leq k \leq K$ on an event with 
	probability at least $1 - \delta/3$.
\end{crllr}
Next, the MSSA procedure comes into the play.
Denote
\[
	\elinei k_m(x) 
	= \frac 1{N_k(x)} \sum\limits_{i =1}^n w_i^{(k)}(X_i, x) \eta_m(X_i),
\]
where $N_k(x) = \sum\limits_{i=1}^n w_i^{(k)}(X_i, x)$,
and for any $\delta \in (0, 1)$ and any $x\in\x$ define
\begin{align}
	\label{kstar}
	\notag
	k^* = k^*(\delta, x) = \max\left\{ k' : |\elinei k_m(x) -\right.
	&
	\elinei {k-1}_m(x)| 
	\leq \sqrt{ \frac{2\log(12KM/\delta)}{u_0 N_k(x)} }
	\\&
	\forall \, 1 \leq m \leq M, \, \forall \, 2 \leq k \leq k' \Big\}.
\end{align}
We call the set $\{ k : 1 \leq k \leq k^*\}$ \emph{the small bias region}.
In this region, MSSA has the following oracle property.
\begin{lmm}
	\label{mssabound}
	Let \eqref{a1} and \eqref{a2} be fulfilled.
	Fix any $\delta \in (0, 1)$ and $x \in \x$ and choose
	\begin{equation}
	\label{z_k}
	z_k = \frac{8 M^2}{u_0} \log \frac{12KM}\delta .	
	\end{equation}
	Then there exists a universal constant $C_1$ {\rm(}depending only on $u_0$ and 
	$u$ from \eqref{a2} {\rm)} such that, with probability at least $1 - 
	\delta/3$ over training samples, it holds
	\[
		|\that_m(x) - \ttildei k_m(x)| \leq C_1 M^{3/2} \sqrt{\frac{\log(12KM/\delta)}{n_k}}
	\]
	simultaneously for all $1 \leq m \leq M$ and $1 \leq k \leq k^*$ with $k^* = k^*(\delta, 
	x)$ given by \eqref{kstar}.
\end{lmm}
The proof of Lemma \ref{mssabound} is given in Appendix \ref{mssaboundproof}.
A natural question arises: how large is the small bias region?
The answer is given in the following lemma.
\begin{lmm}
	\label{kstarbound}
	Assume \eqref{a1}, \eqref{a2} and \eqref{a5}.
	Fix any $x \in \x$ and $\delta \in (0, 1)$.
	Then
	\[
		n_{k^*}
		= n_{k^*(\delta, x)}
		\gtrsim \left( (n\varkappa p(x))^{2\alpha/(2\alpha + d)} \left( 
		\log(12KM/\delta)\right)^{d/(2\alpha + d)} \right) \vee \log(12KM/\delta)
	\]
	with probability at least $(1 - \delta/3)$ over training samples.
\end{lmm}
The proof is moved to Appendix \ref{kstarboundproof}.
We will show later that an optimal value of $n_k$ is less than $n_{k^*}$, so the MSSA 
classifier enjoys a minimax rate of convergence up to a logarithmic factor.

\subsection{Step 2: the case of a bounded away from zero density}
\label{step2}

Corollary \ref{knn_corollary}, Lemma \ref{kstarbound}, and Lemma \ref{mssabound} 
imply that, given $x \in \x$, with probability at least $1 - \delta$, 
simultaneously for all $m$, $1 \leq m \leq M$, and $k \leq k^* = k^*(\delta, x)$ (i. e. $n_k 
\lesssim (n\varkappa p(x))^{2\alpha/(2\alpha + d)} \left( 
\log(12KM/\delta)\right)^{d/(2\alpha + d)} \vee \log(12KM/\delta)$),  it holds
\begin{align}
	\label{eq10}
	|\that_m(x) - \theta_m(x)|
	&
	\leq \frac L{\left(n\varkappa p(x)\right)^{\alpha/d}} \big( 2n_k + 4\log(6KM/\delta) 
	\big)^{\alpha/d}
	\\&\notag
	+ \sqrt{ \frac{\log(12KM/\delta)}{n_k} } + C_1 M^{3/2} 
	\sqrt{\frac{\log(12KM/\delta)}{n_k}} .
\end{align}
Let $r = 2 + \beta$.
Since $|\that_m(x) - \theta_m(x)| \leq 1$ almost surely, the expectation of 
$\max\limits_{1\leq m\leq M} |\that_m(x) - \theta_m(x)|^r$ with respect to training 
samples can be bounded by
\begin{align}
\label{rthmoment}
	&\notag
	\E_{S_n} \max\limits_{1\leq m\leq M} |\that_m(x) - \theta_m(x)|^r
	\\&
	\leq \delta
	+ \min\limits_{1\leq k\leq k^*} \Biggl[ \frac L{\left(n\varkappa p(x)\right)^{\alpha/d}} 
	\big( 2n_k + 4\log(6KM/\delta) \big)^{\alpha/d}
	\\&\notag
	+ \sqrt{ \frac{\log(12KM/\delta)}{n_k} } + C_1 M^{3/2} 
	\sqrt{\frac{\log(12KM/\delta)}{n_k}} \Biggr]^r .
\end{align}
Choose any $k \leq k^*$, fulfilling
\[
	n_k \asymp M^{\frac{3d}{2\alpha +d}} (\varkappa np(x))^{\frac{2\alpha}{2\alpha + d}} 
	\left( \log(12KM/\delta)\right)^{\frac d{2\alpha + d}} .
\]
Existence of such $k$ is guaranteed by Lemma \ref{kstarbound}.
Here and further in this paper, $g(n) \asymp h(n)$ means $g(n) \lesssim h(n) \lesssim 
g(n)$.
Then we have
\begin{equation}
	\label{rthmoment2}
	\E_{S_n} \max\limits_{1\leq m \leq M}|\that_m(x) - \theta_m(x)|^r \lesssim \delta + 
	\left( \frac{M^3 \log(12KM/\delta)}{np(x)} \right)^{\alpha r/(2\alpha + d)} .
\end{equation}
In the case, when there exists $p_0 > 0$, such that $p(x) \geq p_0$ for all $x \in 
\text{supp}(\p_X)$, taking $\delta = \delta_*$ according to \eqref{delta}, we obtain
\begin{equation}
	\label{rthmoment3}
	\E_{S_n} \max\limits_{1\leq m \leq M}|\that_m(x) - \theta_m(x)|^r \lesssim \left( 
	\frac{M^3 \log n}{np_0} \right)^{\alpha r/(2\alpha + d)} .
\end{equation}
Here we used the fact that, due to \eqref{a2}, $K \lesssim \log n$.
Thus, $\log(12KM/\delta) \lesssim \log\log n + \log M + \log n \lesssim \log n$.
The next lemma helps to transform the bound on moments \eqref{rthmoment3} into the 
bound on the mean excess risk $\E_{S_n}(\widehat f)$ of the classifier $\widehat f$.
\begin{lmm}
	\label{fast_rate}
	Let the low noise condition \eqref{a4} be fulfilled.
	Let $\that_m(x)$ be any estimator of $\theta_m(x)$ at the point $x\in \mathcal X$.
	Suppose that for some $r > 1 + \beta$, for all \( m \) from \( 1 \) to \( M \) and for 
	almost all $x$ with respect to $\p_X$, it holds
	\[
		\E_{S_n} \max\limits_{1\leq m\leq M} |\that_m(x) - \theta_m(x)|^r \leq \chi_r,
	\]
	with a function $\chi_r$, which does not depend on $x$.
	Denote a plug-in classifier, associated with the estimates $\that_1(x), \dots, 
	\that_M(x)$, by $\widehat f(x) = \argmax\limits_{1\leq m\leq M} \that_m(x)$.
	Then for the excess risk \( \mathcal E(\widehat f) \) it holds
	\[
		\E_{S_n} \mathcal E(\widehat f) \leq B \left( 1 + \frac{6(r + \beta + 2)}{r - \beta - 
		1} \right)\chi_r^{\frac{1+\beta}r} .
	\]
\end{lmm}
Proof of Lemma \ref{fast_rate} is given in Appendix \ref{fast_rate_proof}.
The inequality \eqref{rthmoment3} and Lemma \ref{fast_rate} immediately yield
\[
	\E_{S_n} \mathcal E(\widehat f) \lesssim \left( \frac{M^3\log n}{np_0} 
	\right)^{\alpha (1+\beta)/(2\alpha + d)},
\]
which finishes the proof of the first part of the bound \eqref{main1}.

\subsection{Step 3: extension to the case of a general density}
\label{step3}

Now, consider a density $p(x)$, which fulfils \eqref{a5} and \eqref{a6}.
Let
\begin{equation}
	\label{pstar}
	p_* = \left( \frac{M^3 \log^2 n}{n} \right)^{\frac{\alpha\beta}{\alpha\beta + 
	p(2\alpha+d)}} .
\end{equation}
Choose $\delta_* = \psi_*^{r_*}$, $r_* = \log \psi_*^{-1}$, and
\[
	\psi_*
	= \left( \frac{M^3 \log^2 n}{n p_*} \right)^{\alpha/(2\alpha+d)}
	= \left( \frac{M^3 \log^2 n}{n} \right)^{\frac{\alpha}{\alpha\beta/p + 2\alpha+d}} .
\]
Define events $B_0 = \{ p(X) \geq p_* \}$ and $B_j = \{ 2^{-j} p_* \leq p(X) < 2^{-j+1} 
p_* \}$, $1 \leq j \leq J$, where
\[
	J = \left\lfloor \frac{\log \psi_*^{-1} - 1}{ \alpha/(2\alpha + d) \log 2 } \right\rfloor .
\]
Note that such choice of $J$ implies
\begin{equation}
	\label{j}
	2^{j\alpha/(2\alpha + d)} \psi_* \leq e^{-1} \quad \forall \, j \leq J.
\end{equation}
Then, using Lemma \ref{aux1}, we have
\begin{align*}
	&
	\E_{S_n} \mathcal E(\widehat f)
	= \E_{S_n} \E_X \left( \eta_{f^*(X)}(X) - \eta_{\widehat f(X)}(X) \right)
	\leq 2\E_{S_n} \E_X \left( \theta_{f^*(X)}(X) - \theta_{\widehat f(X)}(X) \right)
	\\&
	= 2\E_X \E_{S_n} \left[ \theta_{f^*(X)}(X) - \theta_{\widehat f(X)}(X) \right] \II\left( 
	f^*(X) \neq \widehat f(X) \right) \left( \II(B_0) + \sum\limits_{j=1}^J \II(B_j) + \II(B_{J+1})
	\right)
	\\&
	\leq 2\E_X \E_{S_n} \left[ \theta_{f^*(X)}(X) - \theta_{\widehat f(X)}(X) \right] \II\left( 
	f^*(X) \neq \widehat f(X) \right) \II(B_0)
	\\&
	\quad + 2\sum\limits_{j=1}^J \E_X \E_{S_n} \left[ \theta_{f^*(X)}(X) - 
	\theta_{\widehat f(X)}(X) \right] \II\left( f^*(X) \neq \widehat f(X) \right)\II(B_j)
	+ 2\p_X(B_{J+1}).
\end{align*}
For the latter term $\p_X(B_{J+1})$ we simply have
\begin{equation}
	\label{eq11}
	\p_X(B_{J+1}) \leq C 2^{-Jp} p_*^p.
\end{equation}
Consider $\E_X \E_{S_n} \left[ \theta_{f^*(X)}(X) - \theta_{\widehat f(X)}(X) \right] 
\II\left( f^*(X) \neq \widehat f(X) \right) \II(B_0)$.
On $B_0$ we have $p \geq p_*$ and, again, applying the argument we used in the case 
of a bounded away from zero density, we obtain
\begin{equation}
	\label{eq4}
	\E_X \E_{S_n} \left[ \theta_{f^*(X)}(X) - \theta_{\widehat f(X)}(X) \right] 
	\II\left( f^*(X) \neq \widehat f(X) \right) \II(B_0)
	\lesssim \psi_*^{1+\beta}
	= \left( \frac{M^3\log^2 n}{np_*} \right)^{\alpha (1+\beta)/(2\alpha + d)} .
\end{equation}
Now, consider $\E_X \E_{S_n} \left[ \theta_{f^*(X)}(X) - \theta_{\widehat f(X)}(X) 
\right] \II\left( f^*(X) \neq \widehat f(X) \right) \II(B_j)$, $j \in \N$.
Let $\{ t_j : j \in \N \}$ be a sequence of integers, which will be specified later.
Then
\begin{align}
	\label{eq1}
	&\notag
	\E_X \E_{S_n} \left[ \theta_{f^*(X)}(X) - \theta_{\widehat f(X)}(X) \right] 
	\II\left( f^*(X) \neq \widehat f(X) \right) \II(B_j)
	\\&
	= \E_X \E_{S_n} \left[ \theta_{f^*(X)}(X) - \theta_{\widehat f(X)}(X) \right] 
	\II\left( 0 < \theta_{f^*(X)}(X) - \theta_{\widehat f(X)}(X) < 2t_j \right) \II(B_j)
	\\&\notag
	+ \E_X \E_{S_n} \left[ \theta_{f^*(X)}(X) - \theta_{\widehat f(X)}(X) \right] 
	\II\left( \theta_{f^*(X)}(X) - \theta_{\widehat f(X)}(X) \geq 2t_j \right) \II(B_j) .
\end{align}
Due to the tail assumption \eqref{a6}, we have
\begin{align}
	\label{eq2}
	&\notag
	\E_X \E_{S_n} \left[ \theta_{f^*(X)}(X) - \theta_{\widehat f(X)}(X) \right] 
	\II\left( 0 < \theta_{f^*(X)}(X) - \theta_{\widehat f(X)}(X) < 2t_j \right) \II(B_j)
	\\&
	\leq 2t_j \p(B_j)
	\leq 2C t_j \left( 2^{-j+1} p_* \right)^p .
\end{align}
For the second term, again, using the inequality
\begin{align*}
	&
	\theta_{f^*(X)}(X) - \theta_{\widehat f(X)}(X)
	= \left(\theta_{f^*(X)}(X) - \that_{f^*(X)}(X) \right)
	\\&
	+\left(\that_{f^*(X)}(X) - \that_{\widehat f(X)}(X) \right)
	+ \left( \that_{\widehat f(X)}(X) - \theta_{\widehat f(X)}(X) \right)
	\\&
	\leq \left(\theta_{f^*(X)}(X) - \that_{f^*(X)}(X) \right)
	+ \left( \that_{\widehat f(X)}(X) - \theta_{\widehat f(X)}(X) \right)
	\\&
	\leq 2\max\limits_{1\leq m\leq M} |\that_m(X) - \theta_m(X)|,
\end{align*}
one obtains
\begin{align*}
	&
	\E_X \E_{S_n} \left[ \theta_{f^*(X)}(X) - \theta_{\widehat f(X)}(X) \right] 
	\II\left( \theta_{f^*(X)}(X) - \theta_{\widehat f(X)}(X) \geq 2t_j \right) \II(B_j)
	\\&
	\leq \E_X \E_{S_n} \left[ \theta_{f^*(X)}(X) - \theta_{\widehat f(X)}(X) \right] 
	\II\left( \max\limits_{1\leq m\leq M} |\that_m(X) - \theta_m(X)| \geq t_j \right) 
	\II(B_j)
	\\&
	\leq \E_X \E_{S_n} \II\left( \max\limits_{1\leq m\leq M} |\that_m(X) - \theta_m(X)| \geq 
	t_j \right) \II(B_j) .
\end{align*}
With probability at least $1 - \delta_*$ we have
\[
	\max\limits_{1\leq m\leq M} |\that_m(X) - \theta_m(X)|  \II(B_j)
	\lesssim \left( \frac{M^3 \log(12KM/\delta_*)}{2^{-j} n p_*} \right)^{\alpha/(2\alpha + 
	d)}
	\lesssim 2^{j\alpha/(2\alpha + d)} \psi_*.
\]
Here we used that, due to \eqref{a2}, $K \lesssim \log n$, so
\[
	\log(12KM/\delta_*) \lesssim \log\log n + \log M + \log^2 \psi_*^{-1} \lesssim \log^2 n .
\]
Denote $\psi_j = 2^{j\alpha/(2\alpha + d)} \psi_*$.
The Markov inequality yields
\[
	\E_{S_n} \II\left( \max\limits_{1\leq m\leq M} |\that_m(X) - \theta_m(X)| \geq t_j 
	\right) \II(B_j)
	\leq \frac{\delta_* + \psi_j^{r_*}}{t_j^{r_*}}
	= \frac{\psi_*^{r_*} + \psi_j^{r_*}}{t_j^{r_*}}
	\leq \frac{2\psi_j^{r_*}}{t_j^{r_*}},
\]
and therefore,
\begin{equation}
	\label{eq3}
	\E_X \E_{S_n} \left[ \theta_{f^*(X)}(X) - \theta_{\widehat f(X)}(X) \right] 
	\II\left( \theta_{f^*(X)}(X) - \theta_{\widehat f(X)}(X) \geq 2t_j \right) \II(B_j)
	\leq 2C \left(2^{-j+1}p_*\right)^p \frac{\psi_j^{r_*}}{t_j^{r_*}} .
\end{equation}
Thus, taking \eqref{eq1}, \eqref{eq2} and \eqref{eq3} together, one obtains
\begin{equation*}
	\E_X \E_{S_n} \left[ \theta_{f^*(X)}(X) - \theta_{\widehat f(X)}(X) \right] 
	\II\left( f^*(X) \neq \widehat f(X) \right) \II(B_j)
	\leq 2C \left( 2^{-j+1} p_* \right)^p \left( t_j + \frac{\psi_j^{r_*}}{t_j^{r_*}} \right) .
\end{equation*}
Take $t_j = \psi_j^{r_*}$ to balance the two terms.
Then
\begin{equation}
	\label{eq6}
	\E_X \E_{S_n} \left[ \theta_{f^*(X)}(X) - \theta_{\widehat f(X)}(X) \right] 
	\II\left( f^*(X) \neq \widehat f(X) \right) \II(B_j)
	\leq 4C \left( 2^{-j+1} p_* \right)^p \psi_j^{r_*/(r_* + 1)}.
\end{equation}
Note that
\begin{equation}
	\label{eq6'}
	\psi_j^{r_*/(r_* + 1)} = 2^{\frac{\alpha j r_*}{(2\alpha + d)(r_* + 1)}} \psi_*^{r_*/(r_*+1)}
	\leq 2^{\alpha j / (2\alpha + d)} \cdot e \psi_*.
\end{equation}
The last inequality follows from the fact
\[
	\psi_*^{r_*/(r_*+1)}
	= \psi_* \cdot \psi_*^{-1/(r_*+1)}
	= \psi_* e^{\log \psi_*^{-1} /(r_*+1)}
	= \psi_* e^{\log \psi_*^{-1} /(\log\psi_*^{-1}+1)}
	\leq e\psi_*
\]
Inequalities \eqref{eq11}, \eqref{eq4}, \eqref{eq6} and \eqref{eq6'} immediately imply
\begin{align}
\label{eq7}
	&\notag
	\E_{S_n} \mathcal E(\widehat f)
	= \E_X \E_{S_n} \left( \theta_{f^*(X)}(X) - \theta_{\widehat f(X)}(X) \right)
	\lesssim  \psi_*^{1+\beta}
	+ \sum\limits_{j=1}^J \left( 2^{-j+1} p_* \right)^p 2^{\alpha j / (2\alpha + d)} \psi_* + 
	\left( 2^{-J} p_* \right)^p
	\\&
	\leq \psi_*^{1+\beta}
	+ 2^p p_*^p \psi_* \sum\limits_{j=1}^\infty 2^{-j(p - \alpha/(2\alpha+d)} + \left( 2^{-J} 
	p_* \right)^p
	= \psi_*^{1+\beta} + \frac{2^p p_*^p \psi_* \cdot 2^{-p + \alpha/(2\alpha+d)}}{1 - 2^{-p 
	+ \alpha/(2\alpha+d)}} + \left( 2^{-J} p_* \right)^p
	\\&\notag
	= \psi_*^{1+\beta} + \frac{p_*^p \psi_* \cdot 2^{\alpha/(2\alpha+d)}}{1 - 2^{-p + 
	\alpha/(2\alpha+d)}} + \left( 2^{-J} p_* \right)^p
\end{align}
Note that the density level $p_*$, defined by \eqref{pstar}, balances the first and the 
second terms.
We also have $p_* \leq \varepsilon_0$ from \eqref{a6}, provided that $n$ is sufficiently 
large.
Concerning the third term in \eqref{eq7}, we have
\[
	2^{-J-1}p_* \leq (e \psi_*)^{(2\alpha+d)/\alpha} p_* = \frac{eM^3 \log^2 n}{n},
\]
and, since $p > \alpha/(2\alpha + d)$, it holds
\[	
	p > \frac{\alpha(1 + \beta)}{\alpha\beta/p + 2\alpha + d},
\]
and with such choice of $p_*$
\[
	\left(\frac{M^3 \log^2 n}{n} \right)^p
	< \left(\frac{M^3 \log^2 n}{n} \right)^{\frac{\alpha(1 + \beta)}{\alpha\beta/p + 2\alpha 
	+ d}}
	= \left(\frac{M^3 \log^2 n}{n p_*} \right)^{\alpha(1 + \beta)/(2\alpha + d)}
	= \psi_*^{1+\beta},
\]
so the third term in \eqref{eq7} is smaller than the first and the second terms.
Finally,
\[
	\E_{S_n} \mathcal E(\widehat f)
	\lesssim \psi_*^{1+\beta}
	= \left( \frac{M^3\log^2 n}n \right)^{\frac{\alpha(1+\beta)}{\alpha\beta/p + 2\alpha + 
	d}} .
\]

\subsection{Step 4: a bound on the excess risk with high probability}
\label{step4} 

In \eqref{eq10}, taking any $n_k \leq n_{k^*(\delta, X)}$, fulfilling
\[
	n_k \asymp M^{\frac{3d}{2\alpha + d}}(np(X))^{\frac{2\alpha}{2\alpha + d}} \left( 
	\log\frac{12KM}\delta 
	\right)^{\frac d{2\alpha + d}},
\]
we have
\begin{align*}
	\theta_{f^*(X)}(X) - \theta_{\widehat f(X)}(X)
	&
	\leq 2\max\limits_{1\leq m \leq M} |\that_m(X) - \theta_m(X)|
	\\&
	\leq C_2 \left( \frac{M^3 \log(12KM/\delta)}{np(X)} \right)^{\frac\alpha{2\alpha + d}}
\end{align*}
with probability at least $1 - \delta$ over training samples.

Consider an event $\{ p(X) \geq p_* \}$ with $p_* \asymp \left( \frac{\log(12KM/\delta)}n 
\right)^{\frac{\alpha\beta}{\alpha\beta + p(2\alpha + d)}}$.
On this event $f^*(X) \neq \widehat f(X)$ only if $\theta_{(1)}(X) - \theta_{(2)}(X) < 
C_2 \left( \frac{M^3\log(12KM/\delta)}{np_*} \right)^{\frac\alpha{2\alpha + d}}$ 
or if $\max\limits_{1\leq m \leq M} |\that_m(X) - \theta_m(X)| \geq \frac{C_2}2 \left( 
\frac{M^3 \log(12KM/\delta)}{np_*} \right)^{\frac\alpha{2\alpha + d}}$.
This yields,
\begin{align*}
	&
	\p_X\left( f^*(X) \neq \widehat f(X), \, p(X) \geq p_* \right)
	\\&
	\leq \p_X\left( \theta_{(1)}(X) - \theta_{(2)}(X) < C_2 \left( 
	\frac{M^3 \log(12KM/\delta)}{np_*} \right)^{\frac\alpha{2\alpha + d}} \right)
	\\&
	+ \p\left( 
	\max\limits_{1\leq m \leq M} |\that_m(X) - \theta_m(X)| > \frac{C_2}2 \left( 
	\frac{M^3 \log(12KM/\delta)}{np_*} \right)^{\frac\alpha{2\alpha + d}} \right)
	\\&
	\leq \delta + B C_2^\beta \left( \frac{M^3 \log(12KM/\delta)}{np_*} 
	\right)^{\frac{\alpha\beta}{2\alpha + d}}
\end{align*}
Then, the probability of an incorrect prediction can be bounded by
\begin{align*}
	\p_X\left( f^*(X) \neq \widehat f(X) \right)
	&
	= \p_X\left( f^*(X) \neq \widehat f(X) , \, p(X) \geq p_* \right)
	\\&\quad
	+ \p_X\left( f^*(X) \neq \widehat f(X) \, \big| \, p(X) < p_* \right) \p_X\left( p(X) 
	< p_* \right)
	\\&
	\leq \delta + B C_2^\beta \left( \frac{M^3 \log(12KM/\delta)}{np_*} 
	\right)^{\frac{\alpha\beta}{2\alpha + d}} + C p_*^p
	\\&
	\lesssim \delta + \left( \frac{M^3 \log(12KM/\delta)}n 
	\right)^{\frac{\alpha\beta}{\alpha\beta/p + (2\alpha + d)}},
\end{align*}
and it finishes the proof of Theorem \ref{main}.

\section{Auxiliary results}

\begin{lmm}
	\label{kl_properties}
	Denote the Kullback-Leibler divergence between distributions 
	$\text{Bernoulli}(\vartheta)$ and $\text{Bernoulli}(\vartheta')$ by $\kl 
	\vartheta{\vartheta'}$.
	Then for any $\vartheta, \vartheta' \in \left[ 1/(2M), 1 - 1/(2M) \right]$ it holds
	\[
	\frac3M (\vartheta - \vartheta')^2 \leq \kl \vartheta{\vartheta'} \leq M^2 
	(\vartheta - \vartheta')^2 .
	\]
\end{lmm}

\begin{proof}
	The proof relies on some properties of the exponential family of distributions.
	For a random variable $Z \sim \text{Bernoulli}(\vartheta)$ the log-density $\log p(z, 
	\vartheta)$ can be written in the following form
	\[
	\log p(z, \vartheta) = z \log\frac{\vartheta}{1- \vartheta} + \log(1 - \vartheta) .
	\]
	Denote $\nu = \nu(\vartheta) = \log\frac{\vartheta}{1- \vartheta}$ and $D(\nu) = 
	\log(1 + e^\nu)$.
	$\nu$ is called a canonical parameter of the Bernoulli distribution.
	The direct computation shows that
	\begin{align*} 
	&
	\log p(z, \vartheta) = z \nu(\vartheta) - D(\nu(\vartheta)),
	\\&
	\vartheta \equiv D'(\nu(\vartheta)),
	\\&
	\text{Var}(Z) = D''(\nu(\vartheta)),
	\\&
	\kl\vartheta{\vartheta'} = D'(\nu)(\nu - \nu') - D(\nu) + D(\nu') = \frac{D''(\xi)}2 
	(\nu - \nu')^2.
	\end{align*}
	In the last formula we used a notation $\nu = \nu(\vartheta)$, $\nu' = 
	\nu(\vartheta')$ and $\xi$ is a number between $\nu$ and $\nu'$.
	The Lagrange theorem yields
	\[
	\vartheta - \vartheta' = D'(\nu) - D'(\nu') = D''(\zeta)(\nu - \nu'),
	\]
	for some $\zeta$ between $\vartheta$ and $\vartheta'$.
	Thus, we obtain the equality
	\[
	\kl\vartheta{\vartheta'} =  \frac{D''(\xi)}{2 \left(D''(\zeta)\right)^2} (\vartheta - 
	\vartheta')^2,
	\]
	which implies
	\[
	\frac{D_0}{2D_1^2} (\vartheta - \vartheta')^2
	\leq \kl\vartheta{\vartheta'}
	\leq \frac{D_1}{2 D_0^2} (\vartheta - \vartheta')^2
	\]
	with $D_0 = \min\limits_{\xi\in [\nu, \nu']} D''(\xi)$ and $D_1 = \max\limits_{\xi\in [\nu, 
		\nu']} D''(\xi)$.
	
	Now, we use the formula $\text{Var}(Z) = D''(\nu(\vartheta))$ and obtain
	\[
	D''(\nu(\vartheta)) = \vartheta(1 - \vartheta) .
	\]
	If $(\vartheta, \vartheta') \in \left[ 1/(2M), 1 - 1/(2M) \right]^2$ then, taking into 
	account the fact that $M \geq 2$, one has $D_0 = 1/(2M)\big(1 - 1/(2M) \big)$, $D_1 
	= 1/4$ and
	\begin{align*}
	&
	\frac{D_0}{2D_1^2} = \frac4M \left( 1 - \frac1{2M} \right) \geq \frac3M,
	\\&
	\frac{D_1}{2D_0^2} = \frac{M^2}2 \left( 1 - \frac1{2M} \right)^{-2} \leq M^2,
	\end{align*}
	and the proof of Lemma \ref{kl_properties} is finished.
\end{proof} 

\bigskip

\begin{lmm}
	\label{aux1}
	Fix a point $x\in\x$ and denote $m^* \in \argmax\limits_{1\leq m \leq M} \eta_m(x)$.
	Then for any $m \neq m^*$ it holds
	\[
	\eta_{m^*}(x) - \eta_m(x) \leq 2\left( \theta_{m^*}(x) - \theta_m(x) \right) .
	\]
\end{lmm}

\begin{proof}
	There are three cases we have to consider: (i) $\eta_{m^*}(x) > 1 - 1/(2M)$, (ii) 
	$\eta_{m^*}(x) \leq 1 - 1/(2M)$, $\eta_m(x) > 1/(2M)$ and (iii) $\eta_{m^*}(x) \leq 1 - 
	1/(2M)$, $\eta_m(x) \leq 1/(2M)$.
	
	Consider the case (i).
	Note that in this case the condition $\eta_{m^*}(x) > 1 - 1/(2M)$ immediately yields 
	$\eta_m(x) < 1/(2M)$ for all $m \neq m^*$.
	Then $\theta_{m^*}(x) = 1 - 1/(2M)$, $\theta_m(x) = 1/(2M)$ for all $m \neq m^*$ 
	and one has
	\[
	\theta_{m^*}(x) - \theta_m(x)
	= 1 - 1/M
	\geq \frac12
	\geq \frac12\left( \eta_{m^*}(x) - \eta_m(x) \right),
	\]
	where we used $M \geq 2$.
	
	Consider the case (ii). In this case, for all $m\neq m^*$, it holds
	\[
	\eta_{m^*}(x) - \eta_m(x)
	= \theta_{m^*}(x) - \theta_m(x)
	\leq 2 \left( \theta_{m^*}(x) - \theta_m(x) \right) .
	\]
	
	Finally, consider the case (iii).
	Since, $\eta_{m^*}(x) \geq \frac1M$ (otherwise, one gets a contradiction with the 
	fact that $m^* \in \argmax\limits_{1 \leq m \leq M} \eta_m(x)$), it holds
	\[
	\frac{\eta_{m^*}(x) + \eta_m(x)}2 \geq \frac1{2M},
	\]
	and we have for all $m \neq m^*$
	\[
	\theta_{m^*}(x) - \theta_m(x)
	= \eta_{m^*}(x) - \frac1{2M}
	\geq \frac12 \big(\eta_m^*(x) - \eta_m(x) \big) .
	\]
	
\end{proof}

\bigskip

\begin{lmm}
	\label{mleconc}
	Fix a point $x \in \x$, an integer $1 \leq m \leq M$ and a set of weights $\{ w_i(X_i, x) : 
	1 \leq i \leq n\}$.
	Denote
	\begin{align*}
	\etilde_m(x) 
	&= \frac 1{N(x)} \sum\limits_{i =1}^n w_i(X_i, x) \II(Y_i = m),
	\\
	\eline_m(x) 
	&= \frac 1{N(x)} \sum\limits_{i =1}^n w_i(X_i, x) \eta_m(X_i),
	\end{align*}
	where
	\[
	N(x) = \sum\limits_{i=1}^n w_i(X_i, x) .
	\]
	Assume that $0 \leq w_i(x) \leq 1$ for $1 \leq i \leq n$.
	Then, for any $t > 0$, it holds
	\[
	\p_{S_n} \left( \left. |\etilde_m(x) - \eline_m(x)| > t \right| X_1, \dots, X_n \right)
	\leq 2 e^{-2N(x)t^2} .
	\]
\end{lmm}

\begin{proof}
	\begin{align*}
	&
	\p_{S_n} \left( \left. |\etilde_m(x) - \eline_m(x)| > t \right| X_1, \dots, X_n 
	\right)
	\\&
	= \p_{S_n} \left( \big|\sum\limits_{i =1}^n w_i(X_i, x) \left( \II(Y_i = m) - 
	\eta_m(X_i) \right) \big| > N(x)t \, \Big| \, X_1, \dots, X_n \right) .
	\end{align*}
	The Hoeffding inequality yields
	\begin{align*}
	&
	\p_{S_n} \left( \left. |\etilde_m(x) - \eline_m(x)| > t \right| X_1, \dots, X_n 
	\right)
	\leq 2\exp \left\{ - \frac{2N^2(x)t^2}{\sum\limits_{i=1}^n w^2_i(X_i, x)} \right\}
	\\&\qquad
	\leq 2\exp \left\{ - \frac{2N^2(x)t^2}{\sum\limits_{i=1}^n w_i(X_i, x)} \right\}
	= 2 e^{-2N(x)t^2} .
	\end{align*}
\end{proof}

\section{Additional proofs}

\subsection{Proof of Theorem \ref{knnbound}}
\label{knnboundproof}
\begin{proof}
	Fix any $x \in \x$ and $1 \leq m \leq M$.
	Denote
	\[
	\eline^w_m(x) = N_w^{-1}(x) \sum\limits_{i=1}^n w_i(X_i, x) \eta_m(X_i),
	\]
	where
	\[
	N_w(x) = \sum\limits_{i=1}^n w_i(X_i, x) .
	\]
	The triangle inequality yields
	\[
	|\eta_m(x) - \etilde^w_m(x)|
	\leq |\eta_m(x) - \eline^w_m(x)| + |\etilde^w_m(x) - \eline^w_m(x)| .
	\]
	Consider $|\eta_m(x) - \eline^w_m(x)|$.
	Since, according to \eqref{a3}, $\eta_m(\cdot)$ is $(L, \alpha)$-Holder continuous, 
	it holds
	\begin{align*}	
	|\eta_m(x) - \eline^w_m(x)|
	&
	= \left|\eta_m(x) - \frac 1{N_w(x)} \sum\limits_{i : w_i(X_i, x) > 0} w_i(X_i, x) 
	\eta_m(X_i) \right|
	\\&
	\leq \frac 1{N_w(x)} \sum\limits_{i : w_i(X_i, x) > 0} w_i(X_i, x) 
	\left|\eta_m(x) -  \eta_m(X_i) \right|
	\\&
	\leq L \max\limits_{i : w_i(X_i, x) > 0} \| X_i - x \|^\alpha
	= L  \| X_{(\mathsf k)}(x) - x \|^\alpha,
	\end{align*}
	where $X_{(\mathsf k)}(x)$ is the $\mathsf k$-th nearest neighbor of $x$.
	The last equality holds, because $w_i(X_i, x) = 0$ if the point $X_i$ is not amongst 
	$n_k$ nearest neighbors of $x$.
	
	For any $t \in (0, r_0]$, it holds
	\begin{align*}
	\p\left( \| X_{(\mathsf k)}(x) - x \| > t \right)
	&
	= \p\left( \sum\limits_{i=1}^n \II( X_i \in B(x, t)) < \mathsf k \right)
	\\&
	\leq \p\left( \text{Binom}(n, \varkappa p(x) t^d) < \mathsf k \right),
	\end{align*}
	where $\text{Binom}(n, \varkappa p(x) t^d)$ stands for the Binomial random variable 
	with parameters $n$ and $\varkappa p(x) t^d$, and the last inequality follows from 
	\eqref{a5}.
	Next, the Bernstein's inequality yields	
	\begin{align*}
	&
	\p\left( \| X_{(\mathsf k)}(x) - x \| > t \right)
	\\&
	\leq \exp\left\{-\frac{(n\varkappa p(x) t^d - \mathsf k)^2}{2n\varkappa p(x)t^d (1 
		- \varkappa p(x) t^d) + 2(n\varkappa p(x) t^d - n_k)/3}\right\}
	\\&
	\leq \exp\left\{-\frac{(n\varkappa p(x) t^d - \mathsf k)^2}{2n\varkappa p(x)t^d + 
		2(n\varkappa p(x) t^d - \mathsf k)/3}\right\},
	\end{align*}
	provided that $n\varkappa p(x) t^d > \mathsf k$.
	Denote $u = n\varkappa p(x) t^d - \mathsf k$.
	We want to choose $u > 0$ in such a way that
	\[
	\exp\left\{-\frac{u^2}{2(u + \mathsf k) + 2u/3}\right\} \leq \frac\delta2 .
	\]
	This is equivalent to choose $u > 0$ satisfying
	\begin{equation}
	\label{eq8}
	u^2 \geq \frac{8u}3 \log\frac2\delta + 2\mathsf k\log\frac2\delta .
	\end{equation}
	Take $u = \mathsf k + 4\log(2/\delta)$.
	The chain of inequalities
	\begin{align*}
	&
	\mathsf k + 4\log\frac2\delta
	\geq \frac83 \log\frac2\delta + \mathsf k + \frac12\log\frac2\delta
	\\&
	\geq \frac83 \log\frac2\delta + \sqrt{2\mathsf k\log\frac2\delta}
	\geq \frac43 \log\frac2\delta + \sqrt{\left( \frac43 \log\frac2\delta  \right)^2 + 
		2\mathsf k\log\frac2\delta}
	\end{align*}
	ensures that such choice of $u$ fulfils \eqref{eq8}.
	Thus, the choice
	\[
	t^d = \frac1{n\varkappa p(x)} \left( 2\mathsf k + 4\log(2/\delta) \right)
	\]
	yields
	\[
	\exp\left\{-\frac{(n\varkappa p(x) t^d - \mathsf k)^2}{2n\varkappa p(x)t^d + 
		2(n\varkappa p(x) t^d - \mathsf k)/3}\right\} \leq \frac\delta2 .
	\]
	Thus, we proved that, on an event with probability at least $1 - \delta/2$, it holds
	\[
	\| X_{(\mathsf k)}(x) - x \| \leq \frac1{n\varkappa p(x)} \left( 2\mathsf k + 
	4\log(2/\delta) \right) .
	\]
	
	It remains to bound $|\etilde^w_m(x) - \eline^w_m(x)|$.
	Lemma \ref{mleconc} implies
	\[
	\p^{\otimes n} \left( \left. |\etilde^w_m(x) - \eline^w_m(x)| > s \right| X_1, \dots, 
	X_n\right)
	\leq 2e^{-2 N_k s^2}
	\leq 2e^{-n_k s^2} .
	\]
	Then, taking the expectation with respect to $X_1, \dots, X_n$, one obtains
	\[
	\p^{\otimes n} \left( |\etilde^w_m(x) - \eline^w_m(x)| > s \right)
	\leq 2e^{-n_k s^2} .
	\]
	Bringing the two bounds together, one obtains that, with probability at least $1 - 
	\delta$ over training samples, it holds
	\begin{align*}
	|\eta_m(x) - \etilde^w_m(x)|
	&
	\leq L  \| X_{(n_k)}(x) - x \|^\alpha + |\etilde^w_m(x) - \eline^w_m(x)|
	\\&
	\leq \frac L{\left(n\varkappa p(x)\right)^{\alpha/d}} \big( 2\mathsf k + 
	4\log(2/\delta) \big)^{\alpha/d} + \sqrt{ \frac{\log(4/\delta)}{\mathsf k} }.
	\end{align*}
\end{proof}

\subsection{Proof of Lemma \ref{mssabound}}
\label{mssaboundproof}

\begin{proof}
	Equations \eqref{kstar} and \eqref{z_k} and 1-Lipschitzness of the funciton 
	$\varphi(\cdot)$ imply
	\[
	M|\tlinei k_m(x) - \tlinei {k-1}_m(x) |
	\leq \sqrt{\frac{z_k}{4N_k(x)}}, \quad 1 \leq m \leq M, 1 \leq k \leq k^*,
	\]
	where $\tlinei k_m(x) = \varphi\big( \elinei k_m(x) \big)$ and $\tlinei {k-1}_m(x) = 
	\varphi\big( \elinei {k-1}_m(x) \big)$.
	Next, we need an auxiliary result, which is formulated in Lemma \ref{kl_properties}.
	It claims that
	\[
	\sqrt{\frac3M} |\ttildei k_m(x) - \ttildei {k-1}_m(x) |
	\leq \klhalf{ \ttildei k_m(x) }{ \ttildei {k-1}_m(x) }
	\leq M|\ttildei k_m(x) - \ttildei {k-1}_m(x) | .
	\]
	Then for any fixed $1 \leq m \leq M$ and $1 \leq k \leq k^*$ it holds	
	\begin{align*}
	&
	\p_{S_n} \left( \left. N_k(x) \kl{ \ttildei k_m(x) }{ \ttildei {k-1}_m(x) } > 
	z_k \right| X_1, \dots, X_n\right)
	\\&
	= \p_{S_n} \left( \left. \klhalf{ \ttildei k_m(x) }{ \ttildei {k-1}_m(x) } > 
	\sqrt{\frac{z_k}{N_k(x)}} \right| X_1, \dots, X_n\right)
	\\&
	\leq \p_{S_n} \left( \left. M |\ttildei k_m(x) - \ttildei {k-1}_m(x) | > 
	\sqrt{\frac{z_k}{N_k(x)}} \right| X_1, \dots, X_n\right)
	\\&
	\leq \p_{S_n} \left( M |\ttildei k_m(x) - \tlinei k_m(x) | + M|\tlinei k_m(x) - 
	\tlinei {k-1}_m(x) | \right.
	\\& \qquad
	\left. \left. + M|\ttildei {k-1}_m(x) - \tlinei {k-1}_m(x)| > 
	\sqrt{\frac{z_k}{N_k(x)}} \right| X_1, \dots, X_n\right)
	\\&
	\leq \p_{S_n} \left( M |\ttildei k_m(x) - \tlinei k_m(x) | + 
	M|\ttildei{k-1}_m(x) - \tlinei {k-1}_m(x)| \right.
	\\&\qquad
	\left. \left. > \sqrt{\frac{z_k}{4N_k(x)}} \right| X_1, \dots, X_n\right)
	\\&
	\leq \p_{S_n} \left( \left. M |\ttildei k_m(x) - \tlinei k_m(x)| > 
	\sqrt{\frac{z_k}{16N_k(x)}} \right| X_1, \dots, X_n\right)
	\\&\qquad
	+ \p_{S_n} \left( \left. M |\ttildei {k-1}_m(x) - \tlinei {k-1}_m(x) | > 
	\sqrt{\frac{z_k}{16N_k(x)}} \right| X_1, \dots, X_n\right)
	\\&
	\leq \p_{S_n} \left( \left. M |\etildei k_m(x) - \elinei k_m(x)| > 
	\sqrt{\frac{z_k}{16N_k(x)}} \right| X_1, \dots, X_n\right)
	\\&\qquad
	+ \p_{S_n} \left( \left. M |\etildei {k-1}_m(x) - \elinei {k-1}_m(x) | > 
	\sqrt{\frac{u_0 z_k}{16N_{k-1}(x)}} \right| X_1, \dots, X_n\right) .
	\end{align*}
	Lemma \ref{mleconc} yields
	\begin{align*}
	&
	\p_{S_n} \left( \left. M |\etildei k_m(x) - \elinei k_m(x)| > 
	\sqrt{\frac{z_k}{16N_k(x)}} \right| X_1, \dots, X_n\right)
	\leq 2e^{-z_k/(8 M^2)},
	\\&
	\p_{S_n} \left( \left. M |\etildei {k-1}_m(x) - \elinei {k-1}_m(x) | > 
	\sqrt{\frac{u_0 z_k}{16N_{k-1}(x)}} \right| X_1, \dots, X_n\right)\leq 2e^{-u_0 
		z_k/(8 M^2)},
	\end{align*}
	and then
	\begin{align*}
	&
	\p_{S_n} \left( \left. N_k(x) \kl{ \ttildei k_m(x) }{ \ttildei {k-1}_m(x) } > 
	z_k \right| X_1, \dots, X_n\right)
	\\&
	\leq 2e^{-z_k/(8 M^2)} + 2e^{-u_0 z_k/(8 M^2)}
	\leq 4e^{-u_0 z_k/(8 M^2)} .
	\end{align*}
	The union bound implies that the next inequality holds simultaneously for all $1 \leq 
	m \leq M$ and $1 \leq k \leq k^*$:
	\begin{align*}
	\p_{S_n} \left( \left. N_k(x) \kl{ \ttildei k_m(x) }{ \ttildei {k-1}_m(x) } > 
	z_k \right| X_1, \dots, X_n\right)
	\leq 4KMe^{-u_0 z_k/(8 M^2)} .
	\end{align*}
	Now, it is easy to observe that, given $\delta \in (0, 1/3)$, the choice
	\[
	z_k = \frac{8 M^2}{u_0} \log \frac{12KM}\delta
	\]
	ensures that $\thati k_m(x) = \ttildei k_m(x)$ simultaneously for all $m$, $1 \leq m 
	\leq M$, and $k$, $1 \leq k \leq k^*$, with probability at least $1 - \delta/3$ 
	over training samples.
	
	Next, following the proof of Theorem 5.3 in \cite{bs07}, one can easily obtain that 
	it holds almost surely
	\[
	\kl{ \thati k_m(x) }{ \thati{k-1}_m(x) } \leq \frac{z_k}{N_k(x)}, \quad \forall \, 1 \leq 
	m \leq M, \, \forall \, 1 \leq k \leq k^* .
	\]
	This and Lemma \ref{kl_properties} imply
	\[
	| \thati k_m(x) - \thati{k-1}_m(x) | \leq \sqrt{\frac{M z_k}{3N_k(x)}}, \quad \forall \, 
	1 \leq m \leq M, \, \forall \, 1 \leq k \leq k^* .
	\]
	Due to \eqref{a1} and \eqref{a2}, $u_0 \leq N_{k-1}(x)/N_k(x) \leq u$ holds almost 
	surely.
	Then
	\begin{align*}
	|\that_m(x) - \thati k_m(x)|
	&
	\leq \sum\limits_{j = k+1}^K |\thati j_m(x) - \thati {j-1}_m(x)|
	\leq \sum\limits_{j = k+1}^K \sqrt{\frac{M z_j}{3N_j(x)}}
	\\&
	= \frac{2M^{3/2}\sqrt{2\log(12KM/\delta)}}{\sqrt{3u_0}} \sum\limits_{j = k+1}^K 
	\frac1{\sqrt{N_j(x)}}
	\\&
	\leq \frac{2M^{3/2}\sqrt{2\log(12KM/\delta)}}{\sqrt{3u_0}} \sum\limits_{j = 
		1}^{K-k} \sqrt{\frac{u^j}{N_k(x)}}
	\\&
	\leq \frac{2M^{3/2}\sqrt{2\log(12KM/\delta)}}{\sqrt{3u_0}} \sum\limits_{j = 
		1}^{\infty} \sqrt{\frac{u^j}{N_k(x)}}
	\\&
	\leq \frac{2M^{3/2}\sqrt{2\log(12KM/\delta)}}{\sqrt{3u_0}} \cdot \frac{\sqrt{u}}{1 
		- \sqrt{u}} \cdot \frac1{\sqrt{N_k(x)}}
	\\&
	\leq \frac{2M^{3/2}\sqrt{2\log(12KM/\delta)}}{\sqrt{3u_0}} \cdot \frac{\sqrt{u}}{1 
		- \sqrt{u}} \cdot \sqrt{\frac2{n_k}} .
	\end{align*}
	Here we used that, due to \eqref{a1},  $N_k(x) \leq n_k \leq 2N_k(x)$ for any $x$.
	Thus, with probability at least $1 - \delta$ over learning samples, simultaneously for 
	all $1 \leq m \leq M$ and $1 \leq k \leq k^*$ it holds
	\[
	|\that_m(x) - \ttildei k_m(x)| \leq C_1 M^{3/2} \sqrt{\frac{\log(12KM/\delta)}{n_k}}
	\]
	with the constant $C_1 = 4\sqrt{\frac u{3u_0}}(1 - \sqrt u)^{-1}$.
\end{proof}

\subsection{Proof of Lemma \ref{kstarbound}}
\label{kstarboundproof}
\begin{proof}
	Fix some $1 \leq m \leq M$ and $1 \leq k \leq K$. 
	Let $h_{k-1}$ and $h_k$ stand for the distance from $x$ to its $n_{k-1}$-th and 
	$n_k$-th nearest neighbors respectively.
	Then
	\begin{align*}
	| \elinei k_m(x) - \elinei {k-1}_m(x) |
	&
	\leq | \elinei k_m(x) - \eta_m(x)| + | \eta_m(x) - \elinei {k-1}_m(x) |
	\\&
	\leq \sum\limits_{i : \|X_i - x\| \leq h_k} \left| \frac{w_i^{(k)}(X_i, x) 
		\eta_m(X_i)}{N_k(x)} - \eta_m(x) \right|
	\\&
	+ \sum\limits_{i : \|X_i - x\| \leq h_{k-1}} \left| \frac{w_i^{(k-1)}(X_i, x) 
		\eta_m(X_i)}{N_{k-1}(x)} - \eta_m(x) \right|
	\\&
	= \frac 1{N_k(x)} \sum\limits_{i : \|X_i - x\| \leq h_k} \left| w_i^{(k)}(X_i, x) 
	\left(\eta_m(X_i) - \eta_m(x) \right) \right|
	\\&
	+ \frac 1{N_{k-1}(x)} \sum\limits_{i : \|X_i - x\| \leq h_{k-1}} \left| w_i^{(k-1)}(X_i, x) 
	\left(\eta_m(X_i) - \eta_m(x) \right) \right|
	\\&
	\leq \frac 1{N_k(x)} \sum\limits_{i : \|X_i - x\| \leq h_k} w_i^{(k)}(X_i, x) 
	\left|\eta_m(X_i) - \eta_m(x) \right|
	\\&
	+ \frac 1{N_{k-1}(x)} \sum\limits_{i : \|X_i - x\| \leq h_{k-1}} w_i^{(k-1)}(X_i, x) 
	\left|\eta_m(X_i) - \eta_m(x) \right| .
	\end{align*}
	Since $\eta_m(x)$ is $(L, \alpha)$-Holder function, the last expression does not 
	exceed
	\begin{align*}
	&
	\frac 1{N_k(x)} \sum\limits_{i : \|X_i - x\| \leq h_k} w_i^{(k)}(X_i, x) L \|X_i - 
	x\|^\alpha
	\\&
	+ \frac 1{N_{k-1}(x)} \sum\limits_{i : \|X_i - x\| \leq h_{k-1}} 
	w_i^{(k-1)}(X_i, x) L \|X_i - x\|^\alpha
	\\&
	\leq L h_k^\alpha + L h_{k-1}^\alpha
	\leq 2L h_k^\alpha .
	\end{align*}
	Thus,
	\[
	| \elinei k_m(x) - \elinei {k-1}_m(x) | \leq 2L h_k^\alpha .
	\]	
	For any $t \in (0, r_0]$ it holds
	\[
	\p( h_k > t )
	= \p\left( \sum\limits_{i=1}^n \II( X_i \in B(x, t)) \leq n_k \right)
	\leq \p\left( \text{Binom} (n, \varkappa p(x) t^d) \leq n_k \right),
	\]
	where $\text{Binom}(n, \varkappa p(x) t^d)$ stands for the Binomial random variable 
	with parameters $n$ and $\varkappa p(x) t^d$, and the last inequality holds since 
	the condition \eqref{a5} implies $\p_X(B(x, t)) \geq \varkappa p(x) t^d$.
	Next, the Bernstein's inequality yields	
	\begin{align*}
	\p\left( h_k > t \right)
	&
	\leq \exp\left\{-\frac{(n\varkappa p(x) t^d - n_k)^2}{2n\varkappa p(x)t^d (1 - 
		\varkappa p(x) t^d) + 2(n\varkappa p(x) t^d - n_k)/3}\right\}
	\\&
	\leq \exp\left\{-\frac{(n\varkappa p(x) t^d - n_k)^2}{2n\varkappa p(x)t^d + 
		2(n\varkappa p(x) t^d - n_k)/3}\right\},
	\end{align*}
	provided that $n\varkappa p(x) t^d > n_k$.
	Let $u = n\varkappa p(x) t^d - n_k$.
	We want to choose $u > 0$ such that
	\[
	\exp\left\{-\frac{u^2}{2(u + n_k) + 2u/3}\right\} \leq \frac\delta{3KM} .
	\]
	This is equivalent
	\begin{equation}
	\label{eq9}
	u^2 \geq \frac{8u}3 \log\frac3\delta + 2n_k\log\frac{3KM}\delta .
	\end{equation}
	Take $u = n_k + 4\log({3KM}/\delta)$.
	Again, as in the proof of Theorem \ref{knnbound}, the chain of inequalities
	\begin{align*}
	&
	n_k + 4\log\frac{3KM}\delta
	\geq \frac83 \log\frac{3KM}\delta + n_k + \frac12\log\frac{3KM}\delta
	\\&
	\geq \frac83 \log\frac{3KM}\delta + \sqrt{2n_k\log\frac{3KM}\delta}
	\geq \frac43 \log\frac{3KM}\delta + \sqrt{\left( \frac43 \log\frac{3KM}\delta  
		\right)^2 + 2n_k\log\frac{3KM}\delta}
	\end{align*}
	ensures that such $u$ fulfils \eqref{eq9}.
	Thus, the choice
	\[
	t^d = \frac1{n\varkappa p(x)} \left( 2n_k + 4\log({3KM}/\delta) \right)
	\]
	yields
	\[
	\exp\left\{-\frac{(n\varkappa p(x) t^d - n_k)^2}{2n\varkappa p(x)t^d + 
		2(n\varkappa p(x) t^d - n_k)/3}\right\} \leq \frac\delta{3KM} .
	\]
	Thus, with probability at least $(1 - \delta/(3KM))$ over training samples, one has
	\[
	h_k^d
	\leq \frac1{n\varkappa p(x)} \left( 2n_k + 4\log(3KM/\delta) \right) .
	\]
	and
	\[
	|\elinei k_m - \elinei {k-1}_m| \leq 
	2L \left(n\varkappa p(x) \right)^{-\alpha/d} \left( 2n_k + 4\log(3KM/\delta) 
	\right)^{\alpha/d} .
	\]
	Now, fix any $1 \leq k' \leq K$.
	The union bound implies that, with probability at least $1 - \delta/3$, for all $k$, $1 
	\leq k \leq k'$,  and all $m$, $1 \leq m \leq M$, it holds
	\[
	|\elinei k_m(x) - \elinei {k-1}_m(x)| \leq 
	2L \left(n\varkappa p(x) \right)^{-\alpha/d} \left( 2n_k + 4\log(3KM/\delta) 
	\right)^{\alpha/d} \, .
	\]
	It remains to find values of $n_k$ when
	\[
	2L \left(n\varkappa p(x) \right)^{-\alpha/d} \left( 2n_k + 4\log(3KM/\delta) 
	\right)^{\alpha/d} \leq
	\sqrt{\frac{2\log(12KM/\delta)}{ u_0 n_k}} \, .
	\]	
	Let $n_k = c^2 \cdot \log(12KM/\delta)$ and find such values of $c$ that
	\[
	2L \left(n\varkappa p(x) \right)^{-\alpha/d} (2c^2+4)^{\alpha/d} \left( 
	\log(12KM/\delta) \right)^{\alpha/d} \leq \sqrt{\frac2{u_0}} \cdot \frac1c \, .
	\]
	It is equivalent to
	\[
	c(2c^2+4)^{\alpha/d} \leq
	\frac{\sqrt 2}{2L \sqrt{u_0}} \left( \frac{n\varkappa p(x)}{ 
		\log(12KM/\delta)}\right)^{\alpha/d} \, .
	\]
	Denote $c_1 = c \vee 1$.
	One can easily observe that any $c_1$ fulfilling
	\[
	c_1^{(2\alpha+d)/d} \leq
	\frac{\sqrt 2}{2L \sqrt{u_0}} \left( \frac{n\varkappa p(x)}{ 
		5\log(12KM/\delta)}\right)^{\alpha/d}
	\]
	ensures the previous inequality.
	Thus, we finally obtain that if
	\[
	n_k \lesssim \left( (n\varkappa p(x))^{2\alpha/(2\alpha + d)} \left( 
	\log(12KM/\delta)\right)^{d/(2\alpha + d)} \right) \vee \log(12KM/\delta), \quad 1 
	\leq k \leq k',
	\]
	then
	\[
	|\elinei k_m(x) - \elinei {k-1}_m(x)| \leq
	\sqrt{\frac{8\log(12KM/\delta)}{u_0 n_k}}
	\]
	holds simultaneously for $1 \leq m \leq M$ and $1 \leq k \leq k'$ with probability at 
	least $1 - \delta/3$.
	This yields
	\[
	n_{k^*} \geq n_{k'} \asymp \left( (n\varkappa p(x))^{2\alpha/(2\alpha + d)} \left( 
	\log(12KM/\delta)\right)^{d/(2\alpha + d)} \right) \vee \log(12KM/\delta) \, .
	\]
\end{proof}

\subsection{Proof of Lemma \ref{fast_rate}}
\label{fast_rate_proof}

\begin{proof}
	Define $q = 1 + \frac1{r+\beta+2}$.
	Fix an arbitrary $t > 0$ and denote
	\[
	A_i = \left\{ q^{i-1} t < \eta_{f^*(X)}(X) - \eta_{\widehat f(X)}(X) \leq 
	q^i t \right\}, \quad i \geq 1 .
	\]
	Then, due to \eqref{risk} and Lemma \ref{aux1}, we have
	\begin{align*}
	\E_{S_n} \mathcal E(\widehat f)
	&
	= \E_{S_n} R(\widehat f) - R(f^*)
	\\& =
	\E_{S_n} \E_X \left[ \eta_{f^*(X)}(X) - \eta_{\widehat f(X)}(X) \right]
	\\& =
	\E_{S_n} \E_X \left[ \eta_{f^*(X)}(X) - \eta_{\widehat f(X)}(X) \right]\II \left(f^*(X) 
	\neq \widehat f(X) \right)
	\\&
	=
	\E_{S_n} \E_X \left[ \eta_{f^*(X)}(X) - \eta_{\widehat f(X)}(X)\right]  
	\II \left( 0 < \eta_{f^*(X)}(X) - \eta_{\widehat f(X)}(X) \leq t \right)
	\\&
	\quad +
	\sum\limits_{i=1}^\infty \E_{S_n} \E_X \left[ \eta_{f^*(X)}(X) - \eta_{\widehat 
		f(X)}(X)\right] \II (f^*(X) \neq \widehat f(X)) \II(A_i)
	\\ &
	\leq
	t \E_{S_n} \p_X\left( 0 < \eta_{f^*(X)}(X) - \eta_{\widehat f(X)}(X) \leq t \right)
	\\ &
	\quad+
	2\sum\limits_{i=1}^\infty \E_{S_n} \E_X \left[ \theta_{f^*(X)}(X) - \theta_{\widehat 
		f(X)}(X)\right]  \II (f^*(X) \neq \widehat f(X)) \II(A_i)		
	\\&
	\leq
	t \p_X\left( \eta_{(1)}(X) - \eta_{(2)}(X) \leq t \right)
	\\&
	\quad+ 2\sum\limits_{i=1}^\infty q^i t \E_{S_n} \E_X \left[ \II (f^*(X) \neq \widehat 
	f(X)) \II(A_i) \right]
	\\&
	\leq
	Bt^{1 + \beta} + 2\sum\limits_{i=1}^\infty q^i t \E_{S_n} \E_X \left[ \II (f^*(X) \neq 
	\widehat f(X)) \II(A_i) \right] .
	\end{align*}
	Note that $\widehat f(X) \neq f^*(X)$ if and only if $\that_{\widehat f(X)}(X) \geq 
	\that_{f^*(X)}(X)$.
	Then
	\begin{align*}
	\theta_{f^*(X)}(X)
	&\leq
	\that_{f^*(X)}(X)  + |\that_{f^*(X)}(X) - \theta_{f^*(X)}(X)|
	\\ &
	\leq
	\that_{\widehat f(X)}(X) + |\that_{f^*(X)}(X) - \theta_{f^*(X)}(X)|
	\\ &
	\leq
	\theta_{\widehat f(X)}(X) + |\that_{f^*(X)}(X) - \theta_{f^*(X)}(X)| 
	+|\that_{\widehat f(X)}(X) - \theta_{\widehat f(X)}(X)| .
	\end{align*}
	For each $i \in \N$ we have
	\begin{align*}
	\E_{S_n} \E_X \II \Big(f^*(X)
	&
	\neq \widehat f(X)\Big)\II(A_i)
	\leq
	\E_{S_n} \E_X \II \Big( \theta_{f^*(X)}(X) \leq \theta_{\widehat f(X)}(X)
	\\ &
	\quad + |\that_{f^*(X)}(X) - \theta_{f^*(X)}(X)| + |\that_{\widehat f(X)}(X) - 
	\theta_{\widehat f(X)}(X)| \Big) \II(A_i)
	\\ &
	\leq
	\E_{S_n} \E_X \II \Big( |\that_{f^*(X)}(X) - \theta_{f^*(X)}(X)|
	+
	|\that_{\widehat f(X)}(X) - \theta_{\widehat f(X)}(X)| \geq q^{i-1}t \Big) \II(A_i)
	\\ &
	\leq
	\E_{S_n} \E_X \II \Big( |\that_{f^*(X)}(X) - \theta_{f^*(X)}(X)| \geq q^{i-2} t )
	\\ &
	+
	\II (|\that_{\widehat f(X)}(X) - \theta_{\widehat f(X)}(X)| \geq q^{i-2} t \Big) 
	\II(A_i) 
	\\ &
	\leq
	2 \E_{S_n} \E_X \II \left( \max\limits_{1 \leq m \leq M} |\that_m(X) - \theta_m(X)| 
	\geq 
	q^{i-2} t \right)\II(A_i)
	\\ &
	\leq
	2 \E_{S_n} \E_X \II \left( \max\limits_{1 \leq m \leq M} |\that_m(X) - \theta_m(X)| 
	\geq 
	q^{i-2} t \right)\II\left(\eta_{(1)}(X) - \eta_{(2)}(X) \leq q^i t\right)
	\\ &
	=
	2 \E_X \p_{S_n} \left( \max\limits_{1 \leq m \leq M} |\that_m(X) - \theta_m(X)| 
	\geq 
	q^{i-2} t \right)\II\left(\eta_{(1)}(X) - \eta_{(2)}(X) \leq q^i t\right)
	\\ &
	\leq
	2 \E_X \frac{\chi_r}{q^{r(i-2)} t^r} \II\left(\theta_{(1)}(X) - \theta_{(2)}(X) \leq 
	q^i t\right)
	\\ &
	\leq
	\frac{2 \chi_r}{q^{r(i-2)} t^r} \cdot Bq^{\beta i} t^\beta
	=
	2B q^{2r} \cdot \frac{\chi_r}{(q^i t)^{r-\beta}} .
	\end{align*}
	Then
	\begin{align*}
	\E_{S_n} \mathcal E(\widehat f)
	& \leq
	B t^{1+\beta} + \sum\limits_{i=1}^\infty 2Bq^{2r} \cdot 
	\frac{\chi_r}{(q^i t)^{r-\beta-1}}
	\\ &
	=
	B t^{1+\beta}\left(1 + \sum\limits_{i=1}^\infty 2q^{2r} \cdot 
	\frac{\chi_r}{q^{i(r-\beta-1)} t^r} \right)
	\\ &
	=
	B t^{1+\beta}\left( 1 + \sum\limits_{i=1}^\infty 2q^{2r} \cdot 
	\frac{\chi_r}{q^{i(r-\beta-1)} t^r} \right)
	\\ &
	=
	B t^{1+\beta}\left( 1 + \frac{2q^{2r} \chi_r}{(q^{r-\beta-1} - 1) t^r} \right) 
	\\ &
	\leq
	B t^{1+\beta}\left( 1 + \frac{2(r + \beta + 2)q^{r+\beta+1} \chi_r}{(r - \beta - 1) 
		t^r} \right) .
	\end{align*}
	Note that
	\[
	q^{r+\beta+1} = \left( 1 + \frac1{r+\beta+2} \right)^{r+\beta+1} \leq e < 3
	\]
	Now, the choice $t = \chi_r^{1/r}$ implies the assertion of Lemma 
	\ref{fast_rate}.	
\end{proof}

\end{document}